\documentclass[journal]{IEEEtran}
\usepackage{times}
\usepackage{epsfig}
\usepackage{graphicx}
\usepackage{amsmath}
\usepackage{amssymb}
\usepackage{url}

\usepackage{tabularx}
\usepackage{subfigure}
\usepackage{booktabs}
\usepackage{multirow}
\ifCLASSINFOpdf
\else
\fi
\begin{document}
%
\title{Deep Subdomain Adaptation Network for Image Classification}
%
%
%

\author{Yongchun Zhu$^{1,2}$,
        Fuzhen Zhuang$^{1,2}$,
        Jindong Wang$^{3}$,
        Guolin Ke$^{3}$,
        Jingwu Chen$^{4}$,\\
        Jiang Bian$^{3}$,
        Hui Xiong$^{5}$,
        and Qing He$^{1,2}$\\%
        $^1$Key Lab of Intelligent Information Processing of Chinese Academy of Sciences (CAS),\\
        Institute of Computing Technology, CAS, Beijing 100190, China\\
        $^2$University of Chinese Academy of Sciences, Beijing 100049, China \\
        $^3$Microsoft Research, $\quad$
        $^4$ByteDance, $\quad$
        $^5$Rutgers, the State University of New Jersey\\
        \{zhuyongchun18s,zhuangfuzhen,heqing\}@ict.ac.cn, \{jindong.wang,Guolin.Ke,jiang.bian\}@microsoft.com\\
        chenjingwu@bytedance.com, hxiong@rutgers.edu
\thanks{Corresponding author: Fuzhen Zhuang, zhuangfuzhen@ict.ac.cn.}}

\maketitle

\begin{abstract}
For a target task where labeled data is unavailable, domain adaptation can transfer a learner from a different source domain. Previous deep domain adaptation methods mainly learn a global domain shift, i.e., align the global source and target distributions without considering the relationships between two subdomains within the same category of different domains, leading to unsatisfying transfer learning performance without capturing the fine-grained information. 
Recently, more and more researchers pay attention to Subdomain Adaptation which focuses on accurately aligning the distributions of the relevant subdomains. However, most of them are adversarial methods which contain several loss functions and converge slowly. Based on this, we present Deep Subdomain Adaptation Network (DSAN) which learns a transfer network by aligning the relevant subdomain distributions of domain-specific layer activations across different domains based on a local maximum mean discrepancy (LMMD). Our DSAN is very simple but effective which does not need adversarial training and converges fast. The adaptation can be achieved easily with most feed-forward network models by extending them with LMMD loss, which can be trained efficiently via back-propagation. Experiments demonstrate that DSAN can achieve remarkable results on both object recognition tasks and digit classification tasks. Our code will be available at: \url{https://github.com/easezyc/deep-transfer-learning}
\end{abstract}


%
\IEEEpeerreviewmaketitle

\section{Introduction}\label{sec:intro}
In recent years, deep learning methods have achieved impressive success in computer vision~\cite{he2016deep}, which however usually needs large amounts of labeled data to train a good deep network. In the real world, it is often expensive and laborsome to collect enough labeled data. For a target task with the shortage of labeled data, there is a strong motivation to build effective learners that can leverage rich labeled data from a related source domain. However, this learning paradigm suffers from the shift of data distributions across different domains, which will undermine the generalization ability of machine learning models~\cite{pan2010survey,quionero2009dataset}.

\begin{figure*}[t!]
\centering
\begin{minipage}[b]{1\linewidth}
\centering
\includegraphics[width=1\columnwidth]{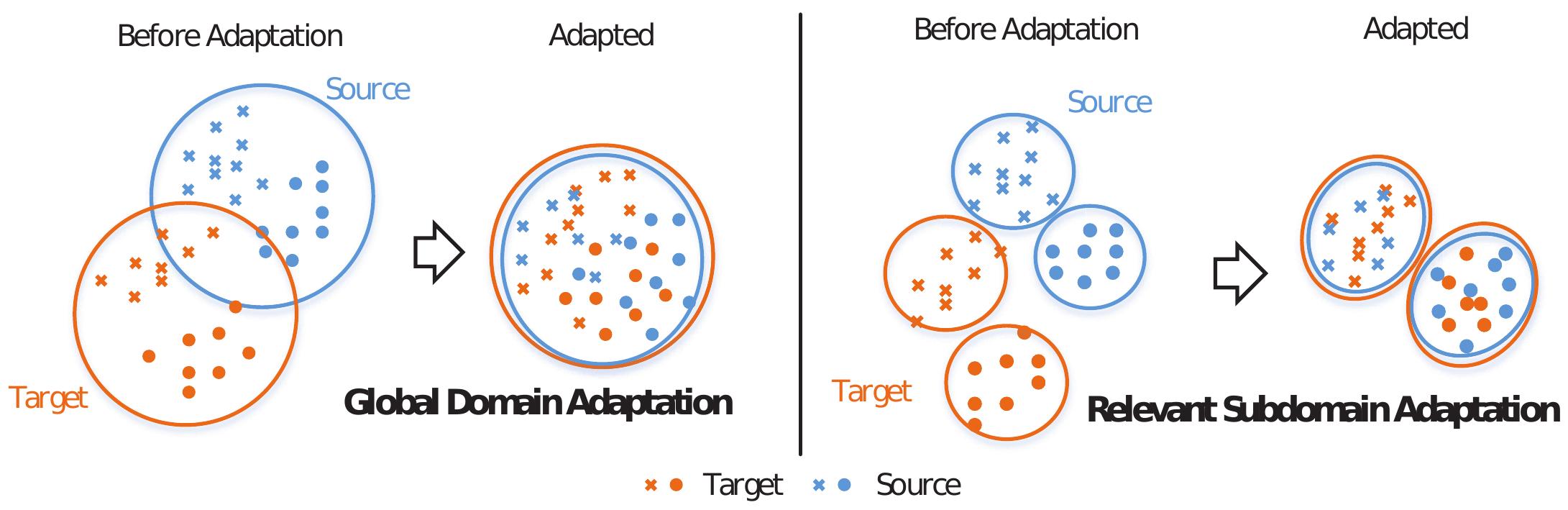}
\end{minipage}
\caption{(left) Global domain adaptation might lose some fine-grained information, (right) while relevant subdomain adaptation can exploit the local affinity to capture the fine-grained information for each category.}\label{fig1}
\end{figure*}

Learning a discriminative model in the presence of the shift between training and test data distributions is known as domain adaptation or transfer learning~\cite{pan2010survey,zhuang2015supervised,zhuang2019comprehensive}. Previous shallow domain adaptation methods bridge the source and target domains by learning invariant feature representations~\cite{gong2012geodesic,long2013transfer,pan2011domain} or estimate instance importance without using target labels~\cite{huang2007correcting}. Recent studies have shown that deep networks can learn more transferable features for domain adaptation~\cite{donahue2014decaf,yosinski2014transferable}, by disentangling explanatory factors of variations behind domains. The latest advantages have been achieved by embedding domain adaptation modules in the pipeline of deep feature learning to extract domain-invariant representations~\cite{ganin2015unsupervised,long2015learning,long2018conditional,pei2018multi,sun2016deep}.

The previous deep domain adaptation methods~\cite{long2015learning,ganin2016domain,sun2016deep} mainly learn a \textit{global domain shift}, i.e., aligning the global source and target distributions without considering the relationships between two subdomains in both domains (A subdomain contains the samples within the same class.). As a result, not only all the data from the source and target domains will be confused, but also the discriminative structures can be mixed up. This might loss the fine-grained information for each category. An intuitive example is shown in Figure~\ref{fig1} (left). After global domain adaptation, the distributions of the two domains are approximately the same, but the data in different subdomains is too close to be classified accurately. This is a common problem in previous global domain adaptation methods. Hence, matching the global source and target domains may not work well for diverse scenarios. 

With regard to the challenge of global domain shift, recently, more and more researchers~\cite{kumar2018co,long2018conditional,pei2018multi,wang2018stratified,xie2018learning,wang2019easy} pay attention to \textit{Subdomain Adaptation} (also called semantic alignment or matching conditional distribution) which is centered on learning a \textit{local domain shift}, i.e., accurately aligning the distribution of the relevant subdomains within the same category in source and target domains. An intuitive example is shown in Figure~\ref{fig1} (right). After \textit{Subdomain Adaptation}, with the local distribution is approximately the same, the global distribution is also approximately the same. However, all of them are adversarial methods which contain several loss functions and converge slowly. We list the comparison of the Subdomain Adaptation methods in Experiment~\ref{experiment}.

Based on the \textit{Subdomain Adaptation}, we propose a Deep Subdomain Adaptation Network (DSAN) to align the relevant subdomain distributions of activations in multiple domain-specific layers across domains for unsupervised domain adaptation. DSAN extends the feature representation ability of deep adaptation networks by aligning relevant subdomain distributions as mentioned above. A key improvement over previous domain adaptation methods is the capability of \textit{Subdomain Adaptation} to capture the fine-grained information for each category, which can be trained in an end-to-end framework. To enable proper alignment, we design a local maximum mean discrepancy (LMMD), which measures the Hilbert-Schmidt norm between kernel mean embedding of empirical distributions of the relevant subdomains in source and target domains with considering the weight of different samples. The LMMD method can be achieved with most feed-forward network models and can be trained efficiently using standard back-propagation. In addition, our DSAN is very simple and easy to implement. Note that the most remarkable results are achieved by adversarial methods recently. Experiments show that DSAN which is a non-adversarial method can obtain the remarkable results for standard domain adaptation on both object recognition tasks and digit classification tasks.

The contributions of this paper are summarized as follows. (1) We propose a novel deep neural network architecture for \textit{Subdomain Adaptation}, which can extend the ability of deep adaptation networks by capturing the fine-grained information for each category. (2) We show that DSAN which is a non-adversarial method can achieve the remarkable results. In addition, our DSAN is very simple and easy to implement.  (3) We propose LMMD to measure the discrepancy between kernel mean embedding relevant subdomains in source and target domains, and successfully apply it to DSAN. (4) A new local distribution discrepancy measure $d_{\mathcal{A}_L}$ is proposed to estimate the discrepancy between two subdomain distributions.
\section{Related Work}\label{sec:relatedWork}
In this section, we will introduce the related work in three aspects: Domain Adaptation, Maximum Mean Discrepancy (MMD) and Subdomain Adaptation methods.

\textbf{Domain Adaptation}. Recent years have witnessed many approaches to solve the visual domain adaptation problem, which is also commonly framed as the visual dataset bias problem~\cite{pan2010survey,quionero2009dataset}. Previous shallow methods for domain adaptation include re-weighting the training data so that they can more closely reflect those in the test distribution~\cite{jiang2007instance}, and finding a transformation in a lower-dimensional manifold that draws the source and target subspaces closer~\cite{gong2012geodesic,long2013transfer,pan2011domain,wang2018visual,wang2019transfer}.

Recent studies have shown that deep networks can learn more transferable features for domain adaptation~\cite{donahue2014decaf,yosinski2014transferable}, by disentangling explanatory factors of variations behind domains. The latest advances have been achieved by embedding domain adaptation modules in the pipeline of deep feature learning to extract domain-invariant representations~\cite{ganin2015unsupervised,long2015learning,long2018conditional,pei2018multi,sun2016deep,zhu2019multi}. Two main approaches are identified among the literature. The first is statistic moment matching based approach, i.e. maximum mean discrepancy (MMD)~\cite{long2015learning,long2016deep,zhu2019aligning}, Central Moment Discrepancy (CMD)~\cite{zellinger2017central}, and second-order statistics matching~\cite{sun2016deep}. The second commonly used approach is based on an adversarial loss, which encourages samples from different domains to be non-discriminative with respect to domain labels, i.e. domain adversarial nets-based adaptation methods~\cite{ganin2016domain,hoffman2017cycada,tzeng2017adversarial} borrowing the idea of GAN. Generally, the adversarial approaches can achieve better performance than the statistic moment matching based approaches. In addition, most state-of-the-art approaches~\cite{long2018conditional,hoffman2017cycada,zhang2018collaborative} are domain adversarial nets-based adaptation methods. Our DSAN is an MMD based method. We show that DSAN without adversarial loss can achieve remarkable results.

\textbf{Maximum Mean Discrepancy}. MMD has been adopted in many approaches~\cite{long2015learning,pan2011domain,zhu2019aligning} for domain adaptation. In addition, there are some extensions of MMD~\cite{long2016deep,long2013transfer}. Conditional MMD~\cite{long2013transfer} and Joint MMD~\cite{long2016deep} measure the Hilbert-Schmidt norm between kernel mean embedding of empirical conditional and joint distributions of the source and target data, respectively. Weighted MMD~\cite{yan2017mind} alleviates class weight bias by assigning class-specific weights to source data. However, our Local MMD measures the discrepancy between kernel mean embedding relevant subdomains in source and target domains with considering the weight of different samples. CMMD~\cite{long2013transfer,wang2017balanced,wang2018visual} is a special case of our LMMD.

\textbf{Subdomain Adaptation}. Recently, we have witnessed considerable interest and research~\cite{kumar2018co,long2018conditional,pei2018multi,xie2018learning} for Subdomain Adaptation which focuses on accurately aligning the distributions of the relevant subdomains. MADA~\cite{pei2018multi} captures multi-mode structures to enable fine-grained alignment of different data distributions based on multiple domain discriminators. MSTN~\cite{xie2018learning} learns semantic representations for unlabeled target samples by aligning labeled source centroid and pseudo-labeled target centroid. CDAN~\cite{long2018conditional} conditions the adversarial adaptation models on discriminative information conveyed in the classifier predictions. Co-DA~\cite{kumar2018co} constructs multiple diverse feature spaces and aligns source and target distributions in each of them individually while encouraging that alignments agree with each other with regard to the class predictions on the unlabeled target examples. The adversarial loss is adopted by all of them. However, compared DSAN with them, our DSAN which is more simple and easy to implement can achieve better performance without adversarial loss. 


\section{Deep Subdomain Adaptation Network}
\label{sec:model}
In unsupervised domain adaptation, we are given a source domain $\mathcal{D}_s=\{(\mathbf{x}^s_i,\mathbf{y}^s_i)\}^{n_s}_{i=1}$ of $n_s$ labeled samples ( $\mathbf{y}^s_i \in \mathbb{R}^C$ is an one-hot vector indicating the label of $\mathbf{x}^s_i$, i.e., $y^s_{ij} = 1$ means $\mathbf{x}^s_i$ belonging to $j$-th class, where $C$ is the number of classes.) and a target domain $\mathcal{D}_t=\{ \mathbf{x}^t_j\}^{n_t}_{j=1}$ of $n_t$ unlabeled samples. $\mathcal{D}_s$ and $\mathcal{D}_t$ are sampled from different data distributions $p$ and $q$ respectively, and $p \neq q$. The goal of deep domain adaptation is to design a deep neural network $\mathbf{y}=f(\mathbf{x})$ that formally reduces the shifts in the distributions of the relevant subdomains in different domains and learns transferable representations simultaneously, such that the target risk $R_t(f)=\mathbb{E}_{(\mathbf{x}, \mathbf{y})~q}[f(\mathbf{x}) \neq \mathbf{y}]$ can be bounded by leveraging the source domain supervised data.

Recent studies reveal that deep networks~\cite{bengio2013representation} can learn more transferable representations than traditional hand-crafted features~\cite{oquab2014learning,yosinski2014transferable}. The favorable transferability of deep features leads to several popular deep transfer learning methods~\cite{long2015learning,long2016deep,ganin2015unsupervised,tzeng2015simultaneous}, which mainly use adaptation layers with a global domain adaptation loss to jointly learn a representation. The formal representation can be:
\begin{equation}
\min_{f} \frac{1}{n_s} \sum^{n_s}_{i=1} J(f( \mathbf{x}^s_i ), \mathbf{y}^s_i) + \lambda \hat{d}(p, q),
\label{preeq}
\end{equation}
where $J(\cdot,\cdot)$ is the cross-entropy loss function (classification loss) and $\hat{d}(\cdot,\cdot)$ is domain adaptation loss. $\lambda > 0$ is the trade-off parameter of the domain adaptation loss and the classification loss.

The common problem with these methods is that they mainly focus on aligning the global source and target distributions without considering the relationships between subdomains within the same category of different domains. These methods derive a \textit{global domain shift} for the source and target domains, and the global distribution of the two domains are approximately the same after adaptation. However, the global alignment may lead to some irrelevant data too close to be classified accurately. Actually, while by exploiting the relationships between subdomains in different domains, just aligning the relevant subdomain distributions cannot only match the global distributions but also the local distributions mentioned above. Therefore, \textit{Subdomain Adaptation} which exploits the relationships between two subdomains to overcome the limitation of aligning global distributions is necessary.

To divide the source and target domains into multiple subdomains which contain the samples within the same class, the relationships between samples should be exploited. It is well known that the samples within the same category are more relevant. However, data in the target domain is unlabeled. Hence we would use the output of the networks as the pseudo labels of target domain data, which will be detailed later. According to the category, we divide $\mathcal{D}_s$ and $\mathcal{D}_t$ into $C$ subdomains $\mathcal{D}_s^{(c)}$ and $\mathcal{D}_t^{(c)}$ where $c \in \{ 1, 2, \dots, C \}$ denotes the class label, and the distributions of $\mathcal{D}_s^{(c)}$ and $\mathcal{D}_t^{(c)}$ are $p^{(c)}$ and $q^{(c)}$ respectively. The aim of \textit{Subdomain Adaptation} is to align the distributions of relevant subdomains which have samples with the same label. Combining the classification loss and subdomain adaptation loss, the loss of \textit{Subdomain Adaptation} method  is formulated as:
\begin{equation}
\min_{f} \frac{1}{n_s} \sum^{n_s}_{i=1} J(f( \mathbf{x}^s_i ), \mathbf{y}^s_i) + \lambda \mathbf{E}_c [ \hat{d}(p^{(c)},q^{(c)}) ],
\label{oureq}
\end{equation}
where $\mathbf{E}_c[\cdot]$ is the mathematical expectation of the class. To compute the discrepancy in Equation~\ref{oureq} between relevant subdomain distributions, based on maximum mean discrepancy (MMD)~\cite{gretton2012kernel} which is a nonparametric measure, we propose Local Maximum Mean Discrepancy to estimate the distribution discrepancy between subdomains.

\subsection{Maximum Mean Discrepancy}
Maximum Mean Discrepancy (MMD)~\cite{gretton2012kernel} is a kernel two-sample test which rejects or accepts the null hypothesis $p=q$ based on the observed samples. The basic idea behind MMD is that if the generating distributions are identical, all the statistics are the same. Formally, MMD defines the following difference measure:
\begin{equation}
d_\mathcal{H}(p,q) \triangleq \| \mathbf{E}_p[\phi (\mathbf{x}^s)]-\mathbf{E}_q[\phi (\mathbf{x}^t)]\|^2_\mathcal{H},
\end{equation}\label{mmd}
where $\mathcal{H}$ is the reproducing kernel Hillbert space (RKHS) endowed with a characteristic kernel $k$. Here $\phi(\cdot)$ denotes some feature map to map the original samples to RKHS and the kernel $k$ means $k(\mathbf{x}^s,\mathbf{x}^t)=\langle \phi (\mathbf{x}^s), \phi (\mathbf{x}^t) \rangle$ where $\langle \cdot , \cdot \rangle$ represents inner product of vectors. The main theoretical result is that $p=q$ if and only if $D_\mathcal{H}(p,q)=0$~\cite{gretton2012kernel}. In practice, an estimate of the MMD compares the square distance between the empirical kernel mean embeddings as
\begin{equation}
\begin{split}
    \hat{d}_\mathcal{H}(p,q)=&\left\| \frac{1}{n_s} \sum_{\mathbf{x}_i\in \mathcal{D}_s }\phi (\mathbf{x}_i)-\frac{1}{n_t} \sum_{\mathbf{x}_j\in \mathcal{D}_t}\phi (\mathbf{x}_j)\right\|^2_\mathcal{H}\\
    =&\frac{1}{n^2_s}\sum^{n_s}_{i=1}\sum^{n_s}_{j=1}k(\mathbf{x}^s_i,\mathbf{x}^s_j)+\frac{1}{n^2_t}\sum^{n_t}_{i=1}\sum^{n_t}_{j=1}k(\mathbf{x}^t_i,\mathbf{x}^t_j)\\
    &-\frac{2}{n_sn_t}\sum^{n_s}_{i=1}\sum^{n_t}_{j=1}k(\mathbf{x}^s_i,\mathbf{x}^t_j),
    \label{unbiased-mmd}
\end{split}
\end{equation}
where $\hat{d}_\mathcal{H}(p,q)$ is an unbiased estimator of $d_\mathcal{H}(p,q)$.

\begin{figure*}[t!]
\centering
\begin{minipage}[b]{1\linewidth}
\centering
\includegraphics[width=.95\linewidth]{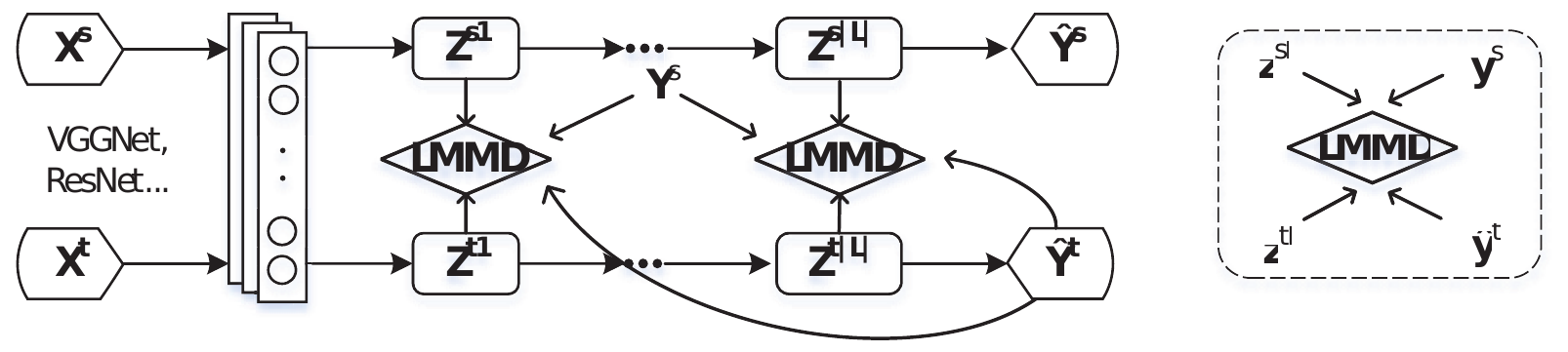}
\end{minipage}
\caption{(left) The architecture of Deep Subdomain Adaptation Network. DSAN will formally reduce the discrepancy between the relevant subdomain distributions of the activations in layers $L$ by using LMMD minimization. (right) The LMMD module needs four inputs: the activations $\mathbf{z}^{sl}$ and $\mathbf{z}^{tl}$ where $l \in L$, the ground-truth label $\mathbf{y}^s$ and the predicted label $\hat{\mathbf{y}}^t$.}\label{fig3} 
\vspace{-0.4cm}
\end{figure*} 

\subsection{Local Maximum Mean Discrepancy}\label{lmmdsec}
As a non-parametric distance estimate between two distributions, MMD has been widely applied to measure the discrepancy between the source and target distributions. Previous deep MMD-based methods~\cite{long2016deep,long2015learning,tzeng2014deep} mainly focus on the alignment of the global distributions, ignoring the relationships between two subdomains within the same category. Taking the relationships of the relevant subdomains into consideration, it is important to align the distributions of the relevant subdomains within the same category in source and target domains. With the desire to align distributions of the relevant subdomains, we propose the Local Maximum Mean Discrepancy (LMMD) as:
\begin{equation}
d_\mathcal{H}(p,q) \triangleq \mathbf{E}_c \| \mathbf{E}_{p^{(c)}}[\phi (\mathbf{x}^{s})]-\mathbf{E}_{q^{(c)}}[\phi (\mathbf{x}^{t})]\|^2_\mathcal{H},
\label{biased-lmmd}
\end{equation}
where $\mathbf{x}^{s}$ and $\mathbf{x}^{t}$ are the instances in $\mathcal{D}_s$ and $\mathcal{D}_t$, and $p^{(c)}$ and $q^{(c)}$ are the distributions of $\mathcal{D}_s^{(c)}$ and $\mathcal{D}_t^{(c)}$ respectively. Different from MMD which focuses on the discrepancy of global distributions, Equation~\ref{biased-lmmd} can measure the discrepancy of local distributions. By minimizing Equation~\ref{biased-lmmd} in deep networks, the distributions of relevant subdomains within the same category are drawn close. Therefore, the fine-grained information is exploited for domain adaptation.

We assume that each sample belongs to each class according to weight $w^c$. Then we formulate an unbiased estimator of Equation~\ref{biased-lmmd} as:
\begin{equation}
\begin{split}
    \hat{d}_\mathcal{H}(p,q)&=\frac{1}{C} \sum^C_{c=1} \left\|  \sum_{\mathbf{x}^s_i\in \mathcal{D}_s} w_i^{sc} \phi (\mathbf{x}^s_i)- \sum_{\mathbf{x}^t_j\in \mathcal{D}_t} w_j^{tc} \phi (\mathbf{x}^t_j)\right\|^2_\mathcal{H},\\
    \label{lmmd}
\end{split}
\end{equation}
where $w_i^{sc}$ and $w_j^{tc}$ denote the weight of $\mathbf{x}^s_i$ and $\mathbf{x}^t_j$ belonging to class $c$. Note that $\sum^{n_s}_{i=1} w_i^{sc}$ and $\sum^{n_t}_{j=1} w_j^{tc}$ are both equal to one, and $\sum_{\mathbf{x}_i\in \mathcal{D}} w_i^c \phi (\mathbf{x}_i)$ is a weighted sum on category $c$. We compute $w_i^c$ for the sample $\mathbf{x}_i$ as:
\begin{equation}
    w_i^c = \frac{ y_{ic} }{ \sum_{ (\mathbf{x}_j,\mathbf{y}_j) \in \mathcal{D} } y_{jc}},
\end{equation}
where $y_{ic}$ is the $c$-th entry of vector $\mathbf{y}_i$. For samples in the source domain, we use the true label $\mathbf{y}_i^s$ as a one-hot vector to compute $w_i^{sc}$ for each sample. However, in unsupervised adaptation where the target domain has no labeled data, we can not calculate Equation~\ref{lmmd} directly with the $\mathbf{y}_j^t$ unavailable. We find that the output of the deep neural network $\mathbf{\hat{y}}_i=f(\mathbf{x}_i)$ is a \textit{probability distribution} which well characterizes the probability of assigning $\mathbf{x}_i$ to each of the $C$ classes. Thus, for target domain $\mathcal{D}_t$ without labels, it is a natural idea to use $\mathbf{\hat{y}}_i^t$ as the probability of assigning $\mathbf{x}^t_i$ to each of the $C$ classes.
Then we can calculate the $w_j^{tc}$ for each target sample. Finally, we can calculate Equation~\ref{lmmd}. 

It is easy to access the labels of the source domain. 
While for the target domain, the label predicted (hard prediction) by the model might be wrong and using this wrong label might degrade the performance. Hence, using the probability prediction (soft prediction) might alleviate the negative impact. 
Note that CMMD which assumes each sample has the same weight is a special case of LMMD, while LMMD takes the uncertainty of target samples into consideration.



To adapt feature layers, we need the activations in the layers. Given source domain $\mathcal{D}_{s}$ with $n_s$ labeled instances and target domain $\mathcal{D}_{t}$ with $n_t$ unlabeled points drawn i.i.d. from $p$ and $q$ respectively, the deep networks will generate activations in layers $l$ as $\{\mathbf{z}_i^{sl}\}^{n_s}_{i=1}$ and $\{\mathbf{z}_j^{tl}\}^{n_t}_{j=1}$. In addition, we cannot compute the $\phi(\cdot)$ directly. Then we reformulate  Equation~\ref{lmmd} as:
\begin{equation}
\begin{split}
    &\hat{d}_l(p,q)=\frac{1}{C}\sum_{c=1}^C  \left [ \sum^{n_s}_{i=1}\sum^{n_s}_{j=1}  w^{sc}_{i} w^{sc}_{j}  k(\mathbf{z}^{sl}_i,\mathbf{z}^{sl}_j) \right .\\
    & + \sum^{n_t}_{i=1}\sum^{n_t}_{j=1}  w^{tc}_{i} w^{tc}_{j} k(\mathbf{z}^{tl}_i,\mathbf{z}^{tl}_j) 
     \left .- 2\sum^{n_s}_{i=1}\sum^{n_t}_{j=1}  w_{i}^{sc} w_{j}^{tc}  k(\mathbf{z}^{sl}_i,\mathbf{z}^{tl}_j)\right ],
    \label{immdfinal}
\end{split}
\end{equation}
where $\mathbf{z}^l$ is the $l$-th ($l \in L=\{ 1, 2, \dots, |L| \}$) layer activation. Equation~\ref{immdfinal} can be used as the adaptation loss in Equation~\ref{oureq} directly, and the LMMD can be achieved with most feed-forward network models.

\subsection{Deep Subdomain Adaptation Network}

Based on LMMD, we propose Deep Subdomain Adaptation Network (DSAN) as shown in Figure~\ref{fig3}. Different from previous global adaptation methods, DSAN not only aligns the global source and target distributions but also aligns the distributions of the relevant subdomains by integrating deep feature learning and feature adaptation in an end-to-end deep learning model. We try to reduce the discrepancy between the relevant subdomain distributions of the activations in layers $L$. We use the LMMD in Equation~\ref{immdfinal} over the domain-specific layers $L$ as the subdomain adaptation loss in Equation~\ref{oureq5},
\begin{equation} \label{oureq5}
\min_{f} \frac{1}{n_s} \sum^{n_s}_{i=1} J(f( \mathbf{x}^s_i ), \mathbf{y}^s_i) + \lambda \sum_{l\in L}\hat{d}_l(p,q).
\end{equation}
Since training deep CNNs requires a large amount of labeled data that is prohibitive for many domain adaptation applications, we start with the CNN models pre-trained on ImageNet 2012 data and fine-tune it as ~\cite{long2016deep}. The training of DSAN mainly follows standard mini-batch stochastic gradient descent (SGD) algorithm. It is worth noting that, with DSAN iteration, the labeling for target samples usually becomes more accurate. This EM-like pseudo label refinement procedure is empirically effective as shown in the experiments.


\textbf{Remark} \label{A-distance}The theory of domain adaptation~\cite{ben2010theory,mansour2009domain} suggests $\mathcal{A}$-distance as a measure of distribution discrepancy, which, together with the source risk, will bound the target risk. The proxy $\mathcal{A}$-distance is defined as $d_\mathcal{A} = 2(1 - 2\epsilon)$, where $\epsilon$ is the generalization error of a classifier (e.g. kernel SVM) trained on the binary problem of discriminating the source and target.
The $\mathcal{A}$-distance just focuses on the global distribution discrepancy, hence we propose $\mathcal{A}_L$-distance to estimate the subdomain distribution discrepancy. First, we define $d_\mathcal{A}$ of class $c$ as $d_{\mathcal{A}_c} = 2(1 - 2\epsilon ^c)$, where $\epsilon ^c$ is the generalization error of a classifier trained on the same class in different domains. Then, we define $d_{\mathcal{A}_L} = \mathbf{E}[d_{\mathcal{A}_c}]=2\mathbf{E}[1 - 2\epsilon ^c]= 2\sum_{c=1}^C p(c)(1 - 2\epsilon ^c)$, where $\mathbf{E}[\cdot]$ denotes Mathematical Expectation and $p(c)$ denotes probability of class $c$ in the target domain.


\subsection{Theoretical Analysis}

In this section, we give an analysis of the effectiveness of using the classifier predictions on the target samples, making use of the theory of domain adaptation~\cite{ben2007analysis,ben2010theory}. 

\textbf{Theorem 1} Let $\mathcal{H}$ be the hypothesis space. Given two domains $\mathcal{S}$ and $\mathcal{T}$, we have
\begin{equation}
    \forall h\in \mathcal{H}, R_\mathcal{T}(h) \leq R_\mathcal{S}(h) + \frac{1}{2}d_{\mathcal{H}\Delta\mathcal{H}}(\mathcal{S}, \mathcal{T})+C,
\end{equation}
where $R_\mathcal{S}(h)$ and $R_\mathcal{T}(h)$ are the expected error on the source samples and target samples, respectively. $R_\mathcal{S}(h)$ can be minimized easily with source label information. Besides, $d_{\mathcal{H}\Delta\mathcal{H}}(\mathcal{S}, \mathcal{T})$ is the domain divergence measure by a discrepancy distance between two distributions $\mathcal{S}$ and $\mathcal{T}$. Actually, there are many approaches to minimize $d_{\mathcal{H}\Delta\mathcal{H}}(\mathcal{S}, \mathcal{T})$, such as adversarial learning~\cite{ganin2015unsupervised}, MMD~\cite{long2015learning}, Coral~\cite{sun2016deep}. $C$ is the shared expected loss and is expected to be negligibly small, thus usually disregarded by previous methods~\cite{long2015learning,ganin2015unsupervised}. However, it is possible that $C$ tends to be large when the cross-domain category alignment is not explicitly enforced. Hence, $C$ needs to be bounded as well. Unfortunately, we cannot directly measure $C$ without target true labels. Therefore, we utilize the pseudo-labels to give the approximate evaluation and minimization.

\textbf{Definition 1} $C$ is defined as:
\begin{equation}
    C = \min_{h \in \mathcal{H}} R_{\mathcal{S}}(h, f_\mathcal{S}) + R_{\mathcal{T}}(h, f_\mathcal{T}),
\end{equation}
where $f_\mathcal{S}$ and $f_\mathcal{T}$ are true labeling functions for source and target domain, respectively.

We show our DSAN is trying to optimize the upper bound for $C$. From~\cite{ben2010theory}, for any labeling functions $f_1, f_2, f_3$, we have:
\begin{equation}
    R(f_1, f_2) \leq R(f_1, f_3) + R(f_2, f_3).
\end{equation}
Then, we have:
\begin{equation}
    \begin{split}
        C =& \min_{h \in \mathcal{H}} R_{\mathcal{S}}(h, f_\mathcal{S}) + R_{\mathcal{T}}(h, f_\mathcal{T})\\
        \leq& \min_{h \in \mathcal{H}} R_{\mathcal{S}}(h, f_\mathcal{S}) + R_{\mathcal{T}}(h, f_\mathcal{S}) + R_{\mathcal{T}}(f_\mathcal{S}, f_\mathcal{T})\\
        \leq& \min_{h \in \mathcal{H}} R_{\mathcal{S}}(h, f_\mathcal{S}) + R_{\mathcal{T}}(h, f_\mathcal{S}) + R_{\mathcal{T}}(f_\mathcal{S}, f_\mathcal{\hat{T}})\\
        &+ R_{\mathcal{T}}(f_\mathcal{T}, f_\mathcal{\hat{T}}),
    \end{split}
\end{equation}
where $f_\mathcal{\hat{T}}$ is pseudo labeling function for target domain. The first term $R_{\mathcal{S}}(h, f_\mathcal{S})$ and the second term $R_{\mathcal{T}}(h, f_\mathcal{S})$ denotes the disagreement between $h$ and the source labeling  function $f_\mathcal{S}$ on source and target samples, respectively. Since $h$ is learned with the labeled source samples, the gap between them can be very small. The last term $R_{\mathcal{T}}(f_\mathcal{T}, f_\mathcal{\hat{T}})$ denotes the discrepancy between the ideal target labeling function $f_\mathcal{T}$ and the pseudo labeling function $f_\mathcal{\hat{T}}$, which would be minimized as learning proceeds. Then, we should focus on the third term $R_{\mathcal{T}}(f_\mathcal{S}, f_\mathcal{\hat{T}})=\mathbf{E}_{x \sim \mathcal{T}}[l(f_\mathcal{S}(x), f_{\mathcal{\hat{T}}}(x))]$, where $l(\cdot, \cdot)$ is typically 0-1 loss function. The source samples of class $k$ would be predicted with label $k$ by the source labeling function $f_\mathcal{S}$. If the feature of target samples in class $k$ is similar with the source feature in class $k$, the target samples can be predicted the same as the pseudo target labeling function. Therefore, if the distributions of subdomains in different domain are matching, $R_{\mathcal{T}}(f_\mathcal{S}, f_\mathcal{\hat{T}})$ is expected to be small.

In summary, by aligning relevant subdomain distributions, our DSAN could further minimize the shared expected loss $C$. Hence, utilizing the prediction of the target samples is effective for unsupervised domain adaptation.

\begin{table*}[!th]
\centering
\caption{Accuracy (\%) on {ImageCLEF-DA} for unsupervised domain adaptation (ResNet50).} \label{tab:Image-CLEF}
\setlength{\tabcolsep}{6mm}{
\begin{tabular}{@{}cccccccc@{}}
\toprule
Method & I $\rightarrow$ P & P $\rightarrow$ I & I $\rightarrow$ C & C $\rightarrow$ I & C $\rightarrow$ P & P $\rightarrow$ C & Avg \\
\midrule
ResNet~\cite{he2016deep} & 74.8$\pm$0.3 & 83.9$\pm$0.1 & 91.5$\pm$0.3 & 78.0$\pm$0.2 & 65.5$\pm$0.3 & 91.2$\pm$0.3 & 80.7\\
DDC~\cite{tzeng2014deep} & 74.6$\pm$0.3 & 85.7$\pm$0.8 & 91.1$\pm$0.3 & 82.3$\pm$0.7 & 68.3$\pm$0.4 & 88.8$\pm$0.2 & 81.8 \\
DAN~\cite{long2015learning} & 75.0$\pm$0.4 & 86.2$\pm$0.2 & 93.3$\pm$0.2 & 84.1$\pm$0.4 & 69.8$\pm$0.4 & 91.3$\pm$0.4 & 83.3\\
DANN~\cite{ganin2016domain} & 75.0$\pm$0.6 & 86.0$\pm$0.3 & 96.2$\pm$0.4 & 87.0$\pm$0.5 & 74.3$\pm$0.5 & 91.5$\pm$0.6 & 85.0\\
D-CORAL~\cite{sun2016deep} & 76.9$\pm$0.2 & 88.5$\pm$0.3 & 93.6$\pm$0.3 & 86.8$\pm$0.6 & 74.0$\pm$0.3 & 91.6$\pm$0.3 & 85.2\\
JAN~\cite{long2016deep} & 76.8$\pm$0.4 & 88.0$\pm$0.2 & 94.7$\pm$0.2 & 89.5$\pm$0.3 & 74.2$\pm$0.3 & 91.7$\pm$0.3 & 85.8\\
MADA~\cite{pei2018multi} & 75.0$\pm$0.3 & 87.9$\pm$0.2 & 96.0$\pm$0.3 & 88.8$\pm$0.3 & 75.2$\pm$0.2 & 92.2$\pm$0.3 & 85.8\\
CAN~\cite{zhang2018collaborative} & 78.2 & 87.5 & 94.2 & 89.5 & 75.8 & 89.2 & 85.7\\
iCAN~\cite{zhang2018collaborative} & 79.5 & 89.7 & 94.7 & 89.9 & 78.5 & 92.0 & 87.4\\
CDAN~\cite{long2018conditional} & 76.7$\pm$0.3 & 90.6$\pm$0.3 & 97.0$\pm$0.4 & 90.5$\pm$0.4 & 74.5$\pm$0.3 & 93.5$\pm$0.4 & 87.1\\
CDAN+E~\cite{long2018conditional} & 77.7$\pm$0.3 & 90.7$\pm$0.2 & \textbf{97.7}$\pm$0.3 & 91.3$\pm$0.3 & 74.2$\pm$0.2 & 94.3$\pm$0.3 & 87.7\\
\midrule
DSAN & \textbf{80.2}$\pm$0.2 & \textbf{93.3}$\pm$0.4 & 97.2$\pm$0.2 & \textbf{93.8}$\pm$0.2 & \textbf{80.8}$\pm$0.4 & \textbf{95.9}$\pm$0.4 & \textbf{90.2}\\
\bottomrule
\end{tabular}
}
\end{table*}
\section{Experiment}\label{experiment}
We evaluate DSAN against competitive transfer learning baselines on object recognition and digit classification. 
The four datasets, including \textbf{ImageCLEF-DA}, \textbf{Office-31}, \textbf{Office-Home}, \textbf{VisDA-2017}, \textbf{Adaptiope} are used for object recognition task, while for digit classification we construct the transfer tasks from \textbf{MNIST}, \textbf{USPS}, \textbf{SVHN}. We denote all transfer tasks as source domain $\rightarrow$ target domain.

\begin{table*}[!th]
\centering
\caption{Accuracy (\%) on Office-31 for unsupervised domain adaptation (ResNet50).}
\setlength{\tabcolsep}{6mm}{
\label{tab:office31}
\begin{tabular}{@{}cccccccc@{}}
\toprule
Method & A $\rightarrow$ W & D $\rightarrow$ W & W $\rightarrow$ D & A $\rightarrow$ D & D $\rightarrow$ A & W $\rightarrow$ A & Avg \\
\midrule
ResNet~\cite{he2016deep} & 68.4$\pm$0.5 & 96.7$\pm$0.5 & 99.3$\pm$0.1 & 68.9$\pm$0.2 & 62.5$\pm$0.3 & 60.7$\pm$0.3 & 76.1\\
DDC~\cite{tzeng2014deep} & 75.8$\pm$0.2 & 95.0$\pm$0.2 & 98.2$\pm$0.1 & 77.5$\pm$0.3 & 67.4$\pm$0.4 & 64.0$\pm$0.5 & 79.7\\
DAN~\cite{long2015learning} & 83.8$\pm$0.4 & 96.8$\pm$0.2 & 99.5$\pm$0.1 & 78.4$\pm$0.2 & 66.7$\pm$0.3 & 62.7$\pm$0.2 & 81.3\\
D-CORAL~\cite{sun2016deep} & 77.7$\pm$0.3 & 97.6$\pm$0.2 & 99.7$\pm$0.1 & 81.1$\pm$0.4 & 64.6$\pm$0.3 & 64.0$\pm$0.4 & 80.8\\
DANN~\cite{ganin2016domain} & 82.0$\pm$0.4 & 96.9$\pm$0.2 & 99.1$\pm$0.1 & 79.7$\pm$0.4 & 68.2$\pm$0.4 & 67.4$\pm$0.5 & 82.2\\
ADDA~\cite{tzeng2017adversarial} & 86.2$\pm$0.5 & 96.2$\pm$0.3 & 98.4$\pm$0.3 & 77.8$\pm$0.3 & 69.5$\pm$0.4 & 68.9$\pm$0.5 & 82.9\\
JAN~\cite{long2016deep} & 85.4$\pm$0.3 & 97.4$\pm$0.2 & 99.8$\pm$0.2 & 84.7$\pm$0.3 & 68.6$\pm$0.3 & 70.0$\pm$0.4 & 84.3\\
MADA~\cite{pei2018multi} & 90.0$\pm$0.1 & 97.4$\pm$0.1 & 99.6$\pm$0.1 & 87.8$\pm$0.2 & 70.3$\pm$0.3 & 66.4$\pm$0.3 & 85.2\\
GTA~\cite{sankaranarayanan2018generate} & 89.5$\pm$0.5 & 97.9$\pm$0.3 & 99.8$\pm$0.4 & 87.7$\pm$0.5 & 72.8$\pm$0.3 & 71.4$\pm$0.4 & 86.6\\
CAN~\cite{zhang2018collaborative} & 81.5 & 98.2 & 99.7 & 85.5 & 65.9 & 63.4 & 82.4\\
iCAN~\cite{zhang2018collaborative} & 92.5 & 98.8 & \textbf{100.0} & 90.1 & 72.1 & 69.9 & 87.2\\
CDAN~\cite{long2018conditional} & 93.1$\pm$0.2 & 98.2$\pm$0.2 & \textbf{100.0}$\pm$.0 & 89.8$\pm$0.3 & 70.1$\pm$0.4 & 68.0$\pm$0.4 & 86.6\\
CDAN+E~\cite{long2018conditional} & \textbf{94.1}$\pm$0.1 & \textbf{98.6}$\pm$0.1 & \textbf{100.0}$\pm$.0 & \textbf{92.9}$\pm$0.2 & 71.0$\pm$0.3 & 69.3$\pm$0.3 & 87.7\\
\midrule
DSAN & 93.6$\pm$0.2 & 98.3$\pm$0.1 & \textbf{100.0}$\pm$0.0 & 90.2$\pm$0.7 & \textbf{73.5}$\pm$0.5 & \textbf{74.8}$\pm$0.4 & \textbf{88.4}\\
\bottomrule
\end{tabular}
}
\end{table*}

\begin{table*}[!th]
\centering
\caption{Accuracy (\%) on {Office-Home} for unsupervised domain adaptation (ResNet50).} \label{tab:officehome}
\begin{tabularx}{\textwidth}{cXXXXXXXXXXXXX}
\toprule
Method & A$\rightarrow$C & A$\rightarrow$P & A$\rightarrow$R & C$\rightarrow$A & C$\rightarrow$P & C$\rightarrow$R & P$\rightarrow$A & P$\rightarrow$C & P$\rightarrow$R & R$\rightarrow$A & R$\rightarrow$C & R$\rightarrow$P & Avg \\
\midrule
ResNet~\cite{he2016deep} & 34.9 & 50.0 & 58.0 & 37.4 & 41.9 & 46.2 & 38.5 & 31.2 & 60.4 & 53.9 & 41.2 & 59.9 & 46.1\\
DAN~\cite{long2015learning} & 43.6 & 57.0 & 67.9 & 45.8 & 56.5 & 60.4 & 44.0 & 43.6 & 67.7 & 63.1 & 51.5 & 74.3 & 56.3\\
DANN~\cite{ganin2016domain} & 45.6 & 59.3 & 70.1 & 47.0 & 58.5 & 60.9 & 46.1 & 43.7 & 68.5 & 63.2 & 51.8 & 76.8 & 57.6\\
JAN~\cite{long2016deep} & 45.9 & 61.2 & 68.9 & 50.4 & 59.7 & 61.0 & 45.8 & 43.4 & 70.3 & 63.9 & 52.4 & 76.8 & 58.3\\
CDAN~\cite{long2018conditional} & 49.0 & 69.3 & 74.5 & 54.4 & 66.0 & 68.4 & 55.6 & 48.3 & 75.9 & 68.4 & 55.4 & 80.5 & 63.8\\
CDAN+E~\cite{long2018conditional} & 50.7 & 70.6 & \textbf{76.0} & 57.6 & \textbf{70.0} & \textbf{70.0} & 57.4 & 50.9 & 77.3 & 70.9 & 56.7 & 81.6 & 65.8\\
\midrule
DSAN & \textbf{54.4} & \textbf{70.8} & 75.4 & \textbf{60.4} & 67.8 & 68.0 & \textbf{62.6} & \textbf{55.9} & \textbf{78.5} & \textbf{73.8} & \textbf{60.6} & \textbf{83.1} & \textbf{67.6} \\
\bottomrule
\end{tabularx}
\end{table*}

\subsection{Setup}
\textbf{ImageCLEF-DA}\footnote{http://imageclef.org/2014/adaptation
} is a benchmark dataset for ImageCLEF 2014 domain adaptation challenge, which is organized by selecting 12 common categories shared by the following three public datasets, each is considered as a domain: Caltech-256 (\textbf{C}), ImageNet ILSVRC 2012 (\textbf{I}), and Pascal VOC 2012 (\textbf{P}). There are 50 images in each category and 600 images in each domain. We use all domain combinations and build 6 transfer tasks: \textbf{I} $\rightarrow$ \textbf{P}, \textbf{P} $\rightarrow$ \textbf{I}, \textbf{I} $\rightarrow$ \textbf{C}, \textbf{C} $\rightarrow$ \textbf{I}, \textbf{C} $\rightarrow$ \textbf{P}, \textbf{P} $\rightarrow$ \textbf{C}.

\begin{table*}[htbp]
  \centering
  \caption{Accuracy (\%) on VisDA-2017 for unsupervised domain adaptation (ResNet101).}
  \setlength{\tabcolsep}{2.5mm}{
    \begin{tabular}{cccccccccccccc}
    \toprule
    Method & airplane & bicycle & bus   & car   & horse & knife & motorcycle & person & plant & skateboard & train & truck & Avg \\
    \midrule
    ResNet~\cite{he2016deep} & 72.3  & 6.1   & 63.4  & \textbf{91.7} & 52.7  & 7.9   & 80.1  & 5.6   & 90.1  & 18.5  & 78.1  & 25.9  & 49.4 \\
    DANN~\cite{ganin2016domain}  & 81.9  & \textbf{77.7} & 82.8  & 44.3  & 81.2  & 29.5  & 65.1  & 28.6  & 51.9  & 54.6  & 82.8  & 7.8   & 57.4 \\
    DAN~\cite{long2015learning}   & 68.1  & 15.4  & 76.5  & 87.0    & 71.1  & 48.9  & 82.3  & 51.5  & 88.7  & 33.2  & 88.9  & 42.2  & 62.8 \\
    JAN~\cite{long2016deep}   & 75.7  & 18.7  & 82.3  & 86.3  & 70.2  & 56.9  & 80.5  & 53.8  & 92.5  & 32.2  & 84.5  & \textbf{54.5} & 65.7 \\
    MCD~\cite{saito2018maximum} & 87.0    & 60.9  & \textbf{83.7} & 64.0    & \textbf{88.9} & \textbf{79.6} & 84.7  & \textbf{76.9} & 88.6  & 40.3  & 83.0 & 25.8  & 71.9 \\
    \midrule
    DSAN  & \textbf{90.9} & 66.9  & 75.7  & 62.4  & \textbf{88.9} & 77.0 & \textbf{93.7} & 75.1  & \textbf{92.8} & \textbf{67.6} & \textbf{89.1} & 39.4  & \textbf{75.1} \\
    \bottomrule
    \end{tabular}%
}
  \label{tab:visda}%
\end{table*}%

\begin{table}[!th]
\centering
\caption{Accuracy (\%) on Adaptiope for unsupervised domain adaptation (ResNet50).}
\setlength{\tabcolsep}{1mm}{
\label{tab:adaptiope}
\begin{tabular}{@{}cccccccc@{}}
\toprule
Method & P$\rightarrow$R & P$\rightarrow$S & R$\rightarrow$P & R$\rightarrow$S & S$\rightarrow$P & S$\rightarrow$R & Avg \\
\midrule
ResNet~\cite{he2016deep} & 63.6 & 26.7 & 85.3 & 27.6 & 7.6 & 2.0 & 35.5\\
RSDA-DANN~\cite{ganin2016domain} & \textbf{78.6} & 48.5 & 90.0 &	43.9 & 63.2 & 37.0 & 60.2\\
RSDA-MSTN~\cite{xie2018learning} & 73.8 & 59.2 & 87.5 & 50.3 & \textbf{69.5} & 44.6 & 64.2\\
\midrule
DSAN & 77.8 & \textbf{60.1} & \textbf{91.9} & \textbf{55.7} & 68.8 & \textbf{47.8} & \textbf{67.0}\\
\bottomrule
\end{tabular}
}
\end{table}

\textbf{Office-31}~\cite{saenko2010adapting} is a benchmark dataset for domain adaptation, comprising 4,110 images in 31 classes collected from three distinct domains: Amazon (\textbf{A}), which contains images downloaded from amazon.com, Webcam (\textbf{W}) and DSLR (\textbf{D}), which contain images taken by web camera and digital SLR camera with different photographical settings, respectively. To enable unbiased evaluation, we evaluate all methods on all 6 transfer tasks \textbf{A} $\rightarrow$ \textbf{W}, \textbf{D} $\rightarrow$ \textbf{W}, \textbf{W} $\rightarrow$ \textbf{D}, \textbf{A} $\rightarrow$ \textbf{D}, \textbf{D} $\rightarrow$ \textbf{A}, \textbf{W} $\rightarrow$ \textbf{A} as in~\cite{long2016deep,tzeng2014deep,ganin2015unsupervised}.

\textbf{Office-Home}~\cite{venkateswara2017deep} is a new dataset, which consists of 15,588 images and is much larger than {Office-31} and {ImageCLEF-DA}. It consists of images from 4 different domains: Artistic images (\textbf{A}), Clip Art (\textbf{C}), Product images (\textbf{P}) and Real-World images (\textbf{R}). For each domain, the dataset contains images of 65 object categories collected in office and home settings. Similarly, we use all domain combinations and construct 12 transfer tasks.

\textbf{VisDA-2017}~\cite{peng2017visda} is a challenging simulation-to-real dataset, with two very distinct domains: \textbf{Synthetic}, renderings of 3D models from different angles and with different lightning conditions; \textbf{Real}, natural images. It contains over 280K images across 12 classes in the training, validation and test domains.

\textbf{Adaptiope}~\cite{ringwald2021adaptiope} is a novel, large scale UDA dataset with 123 classes and a difficult synthetic to real transfer task which provided a new challenge for modern unsupervised domain adaptation algorithms.

\textbf{MNIST-USPS-SVHN}. We explore three digit datasets: MNIST~\cite{lecun1998gradient}, USPS and SVHN~\cite{netzer2011reading} for transfer digit classification. Different from Office-31, MNIST contains grey digit images of size 28$\times$28, USPS contains 16$\times$16 grey digits and SVHN contains color 32$\times$32 digits images which might contain more than one digit in each image. We conduct experiments on three transfer tasks \textbf{MNIST} $\rightarrow$ \textbf{USPS}, \textbf{USPS} $\rightarrow$ \textbf{MNIST}, \textbf{SVHN} $\rightarrow$ \textbf{MNIST}.

\textbf{Baseline Methods} For ImageCLEF-DA and Office-31, we compare our model DSAN with sevaral standard deep learning methods and deep transfer learning methods: Deep Convolutional Neural Network (\textbf{ResNet})~\cite{he2016deep}, Deep Domain Confusion (\textbf{DDC})~\cite{tzeng2014deep}, Deep Adaptation Network (\textbf{DAN})~\cite{long2015learning}, Deep CORAL (\textbf{D-CORAL})~\cite{sun2016deep}, Domain Adversarial Neural Networks (\textbf{DANN})~\cite{ganin2016domain}, Residual Transfer Network (RTN)~\cite{long2016deep},
Adversarial Discriminative Domain Adaptation (ADDA)~\cite{tzeng2017adversarial}, Joint Adaptation Networks (\textbf{JAN})~\cite{long2016deep}, Multi-Adversarial Domain Adaptation (\textbf{MADA})~\cite{pei2018multi}, Collaborative and Adversarial Network (CAN and iCAN)~\cite{zhang2018collaborative}, Generate to Adapt (\textbf{GTA})~\cite{sankaranarayanan2018generate} and Conditional Adversarial Domain Adaptation (\textbf{CDAN} and \textbf{CDAN+E})~\cite{long2018conditional}. For Office-Home, we compare DSAN with ResNet, DAN, DANN, JAN, CDAN, and the results of all baselines are extracted from~\cite{long2018conditional,pei2018multi}. For VisDA-2017, we compare DSAN with ResNet, DANN, DAN, JAN, MCD~\cite{saito2018maximum}, and the results of all baselines are extracted from~\cite{kang2019contrastive}.

\begin{table}[!th]
\centering
\caption{Accuracy (\%) on digit recognition tasks for unsupervised domain adaptation. (`-' means that we did not find the result on the task.)} \label{tab:digit}
\begin{tabular}{@{}cccc@{}}
\toprule
\multirow{2}{*}{Method} & MNIST $\rightarrow$ & USPS $\rightarrow$  & SVHN $\rightarrow$ \\
& USPS &  MNIST & MNIST \\
\midrule
SourceOnly & 75.2$\pm$0.16 & 57.1$\pm$0.17 & 60.1$\pm$0.11 \\
DANN~\cite{ganin2016domain} & 77.1$\pm$1.8 & 73.0$\pm$2.0 & 73.91$\pm$0.07 \\
DRCN~\cite{ghifary2016deep} & 91.8$\pm$0.09 & 73.7$\pm$0.04 & 82.0$\pm$0.16 \\
CoGAN~\cite{liu2016coupled} & 91.2$\pm$0.8 & 89.1$\pm$0.8 & - \\
ADDA~\cite{tzeng2017adversarial} & 89.4$\pm$0.2 & 90.1$\pm$0.8 & 76.0$\pm$1.8 \\
UNIT~\cite{liu2017unsupervised} & 95.97 & 93.58 & 90.53 \\
ATDA~\cite{saito2017asymmetric} & 93.17 & 84.14 & 85.8 \\
GTA~\cite{sankaranarayanan2018generate} & 92.8$\pm$0.9 & 90.8$\pm$1.3 & \textbf{92.4}$\pm$0.9 \\
MSTN~\cite{xie2018learning} & 92.9$\pm$1.1 & - & 91.7$\pm$1.5 \\
\midrule
DSAN & \textbf{96.9}$\pm$0.2 & \textbf{95.3}$\pm$0.1 & 90.1$\pm$0.4\\
\bottomrule
\end{tabular}
\end{table}

For MNIST-USPS-SVHN, we compare DSAN with Domain Adversarial Neural Networks (\textbf{DANN})~\cite{ganin2016domain}, Deep Reconstruction-classification Networks (\textbf{DRCN})~\cite{ghifary2016deep}, Coupled Generative Adversarial Networks (CoGAN)~\cite{liu2016coupled}, Adversarial Discriminative Domain Adaptation (\textbf{ADDA})~\cite{tzeng2017adversarial}, Unsupervised Image-to-image Translation Networks (UNIT)~\cite{}, Asymmetric Tri-training Domain Adaptation (\textbf{ATDA})~\cite{saito2017asymmetric}, Generate to Adapt (\textbf{GTA})~\cite{sankaranarayanan2018generate} and Moving Semantic Transfer Network (\textbf{MSTN})~\cite{xie2018learning}. The results of SourceOnly, DANN, DRCN, CoGAN, ADDA and GTA are extracted from ~\cite{sankaranarayanan2018generate}. For the rest, we refer to the results in their original papers.

\begin{figure*}[t!]
\centering
\subfigure[JAN A$\rightarrow$W]{
\begin{minipage}[b]{0.23\linewidth}
\centering
\includegraphics[width=1.\columnwidth,height=.8\columnwidth]{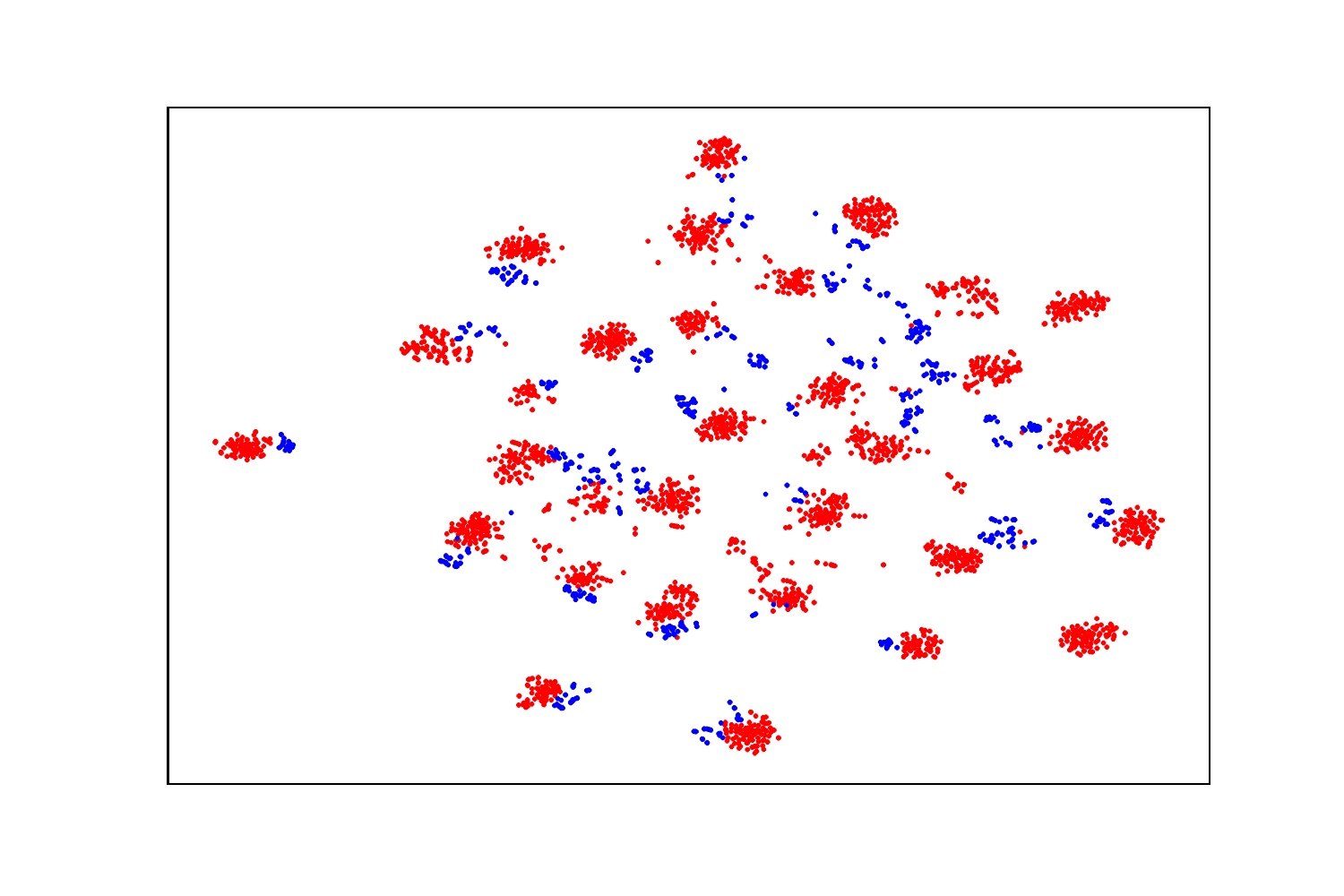}
\label{fig:3a}
\end{minipage}
}
\subfigure[DSAN A$\rightarrow$W]{
\begin{minipage}[b]{0.23\linewidth}
\centering
\includegraphics[width=1.\columnwidth,height=.8\columnwidth]{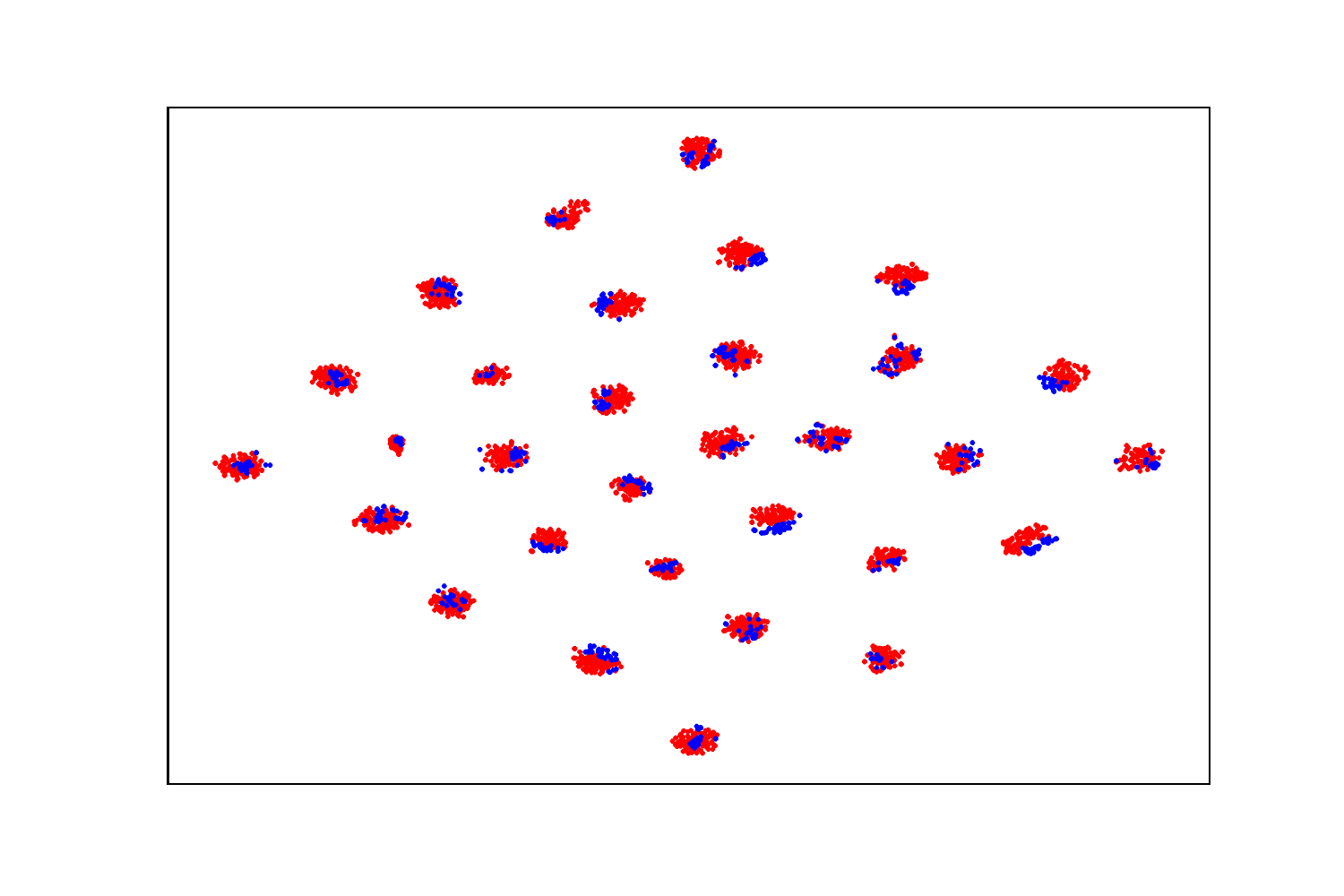}
\label{fig:3b}
\end{minipage}
}
\subfigure[$\mathcal{A}, \mathcal{A}_L$(-distance)]{
\begin{minipage}[b]{0.23\linewidth}
\centering
\includegraphics[width=1.\columnwidth,height=.8\columnwidth]{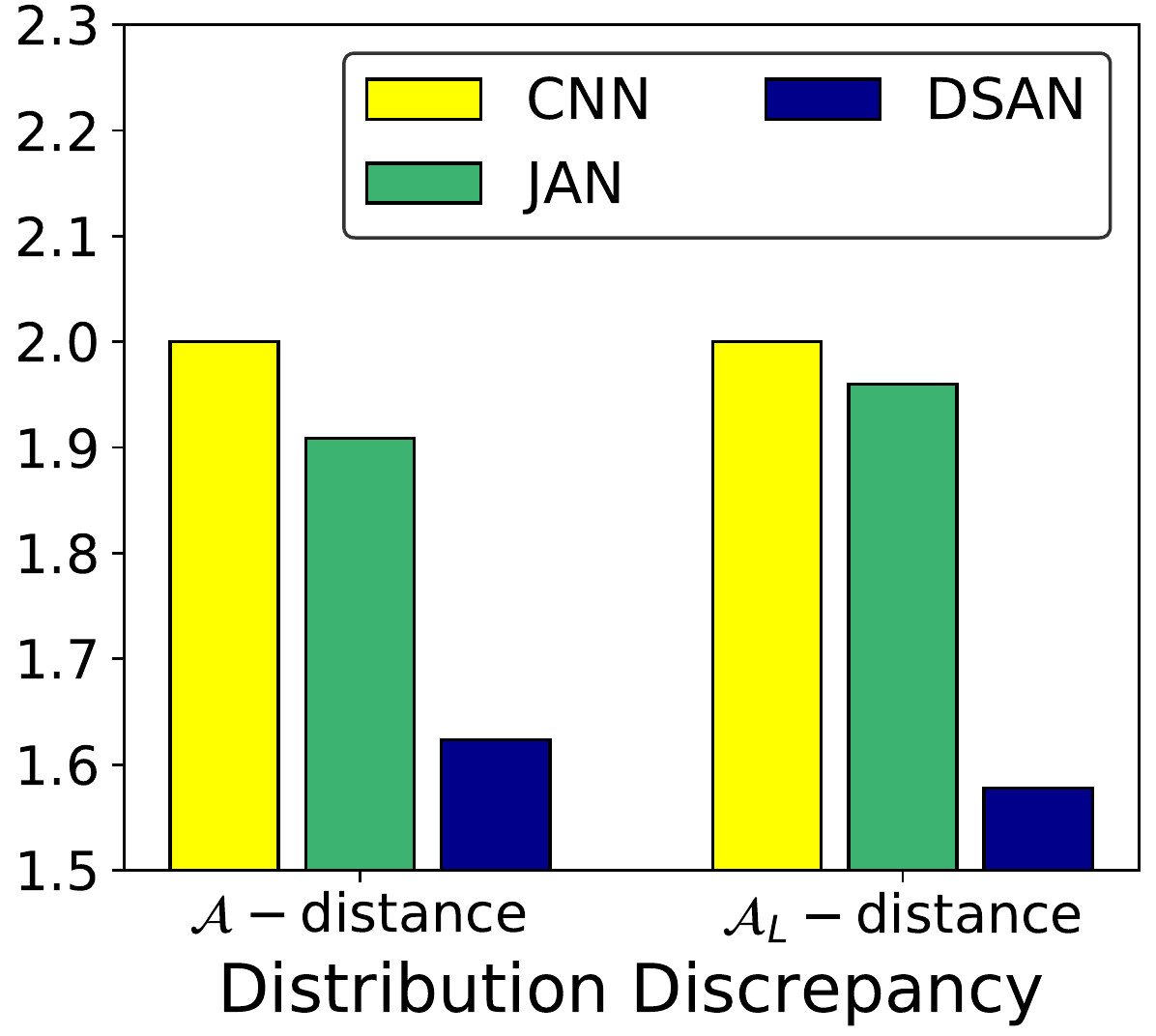}
\label{fig:3c}
\end{minipage}
}
\subfigure[MMD and LMMD]{
\begin{minipage}[b]{0.23\linewidth}
\centering
\includegraphics[width=1.\columnwidth,height=.8\columnwidth]{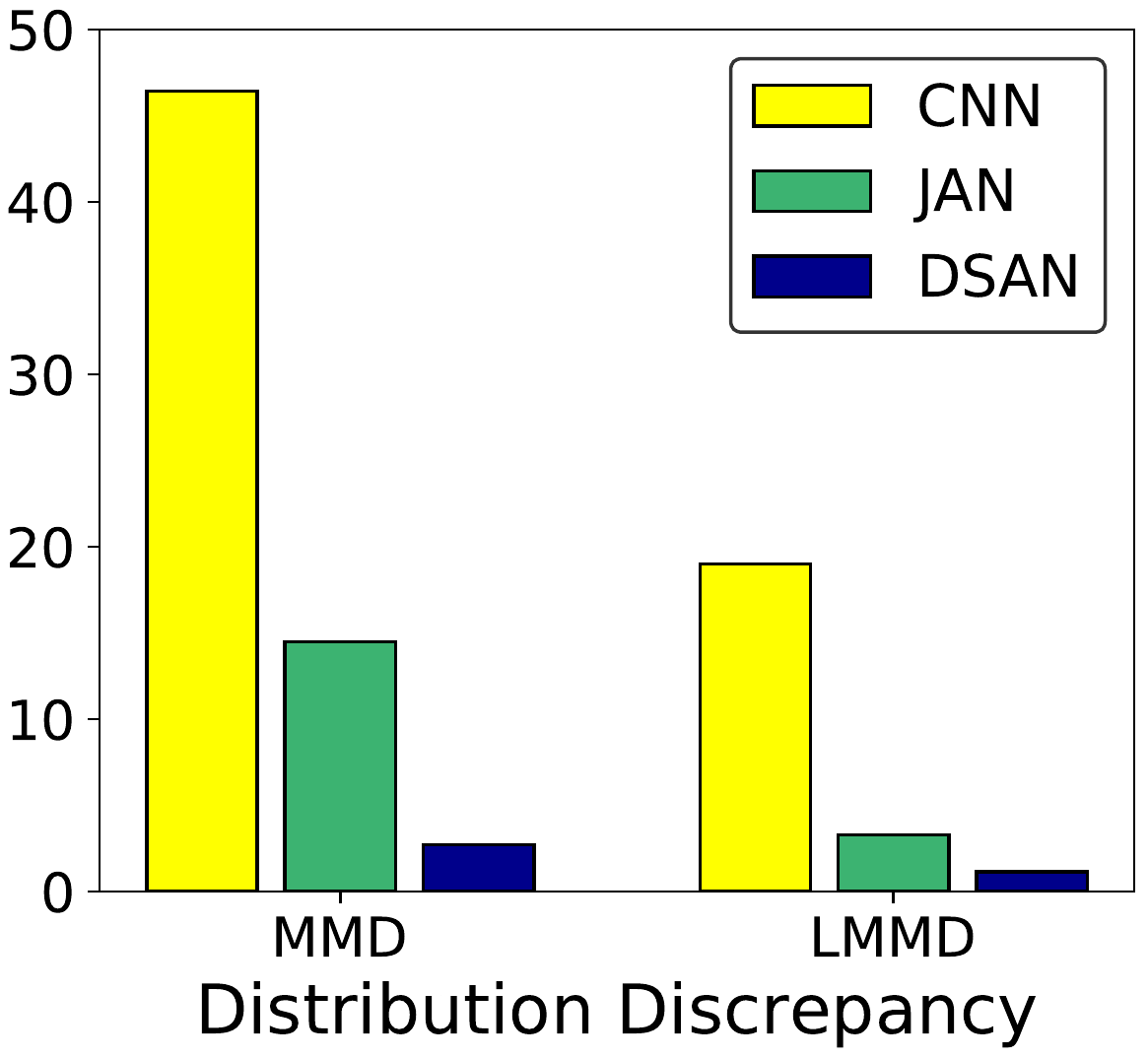}
\label{fig:3d}
\end{minipage}
}
\caption{(a) and (b) are the visualizations of the learned representations using t-SNE for JAN and DSAN on task A$\rightarrow$W, respectively. Red points are source samples and blue are target samples. (c) analyses $\mathcal{A}$-distance and $\mathcal{A}_L$-distance on task A$\rightarrow$W. (d) computes MMD and LMMD  on task A$\rightarrow$W.}
\label{fig:3}
\end{figure*}

\begin{figure}[t!]
\centering
\subfigure[Convergence (iteration)]{
\begin{minipage}[b]{0.46\linewidth}
\centering
\includegraphics[width=1.\columnwidth]{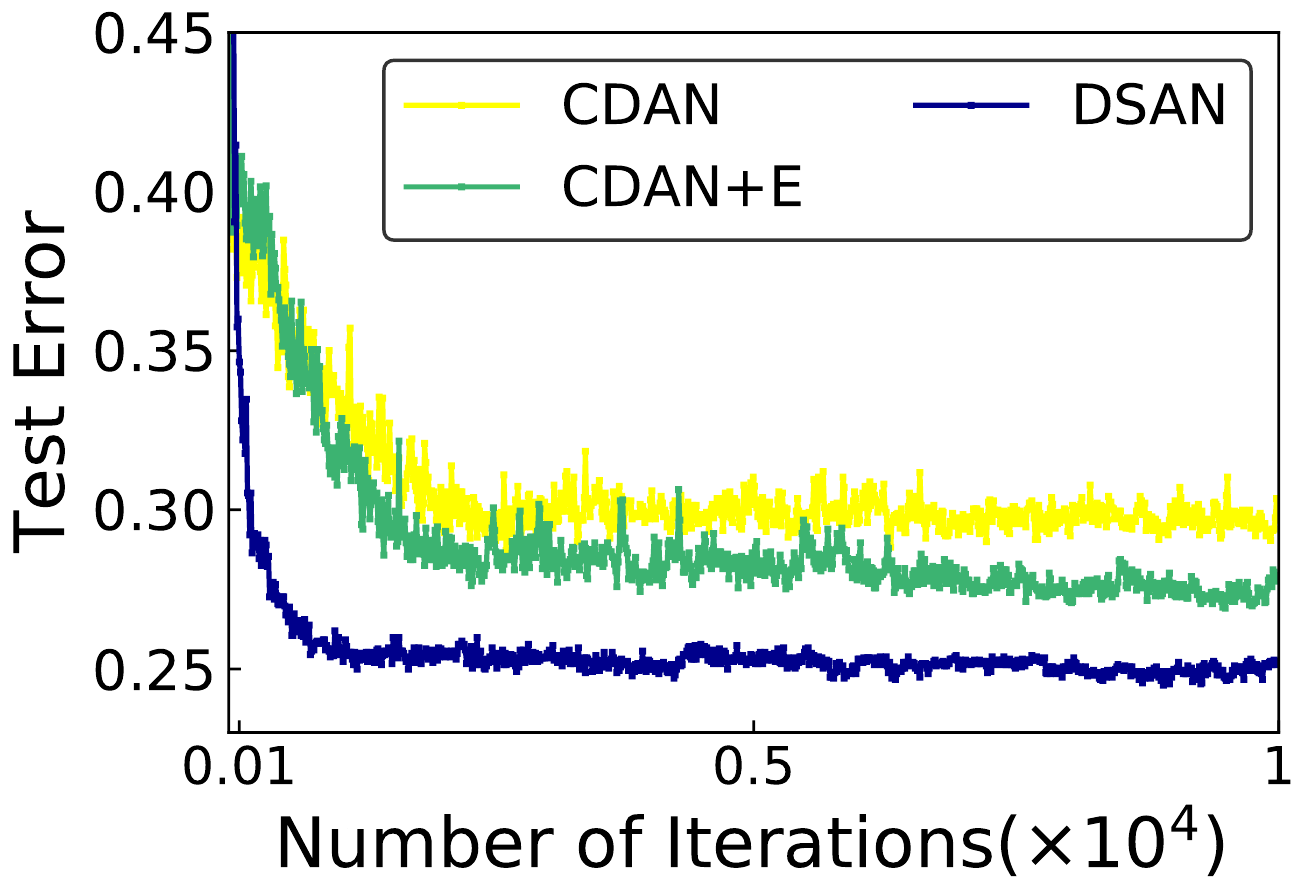}
\label{fig:convergencea}
\end{minipage}
}
\subfigure[Convergence (time)]{
\begin{minipage}[b]{0.46\linewidth}
\centering
\includegraphics[width=1.\columnwidth]{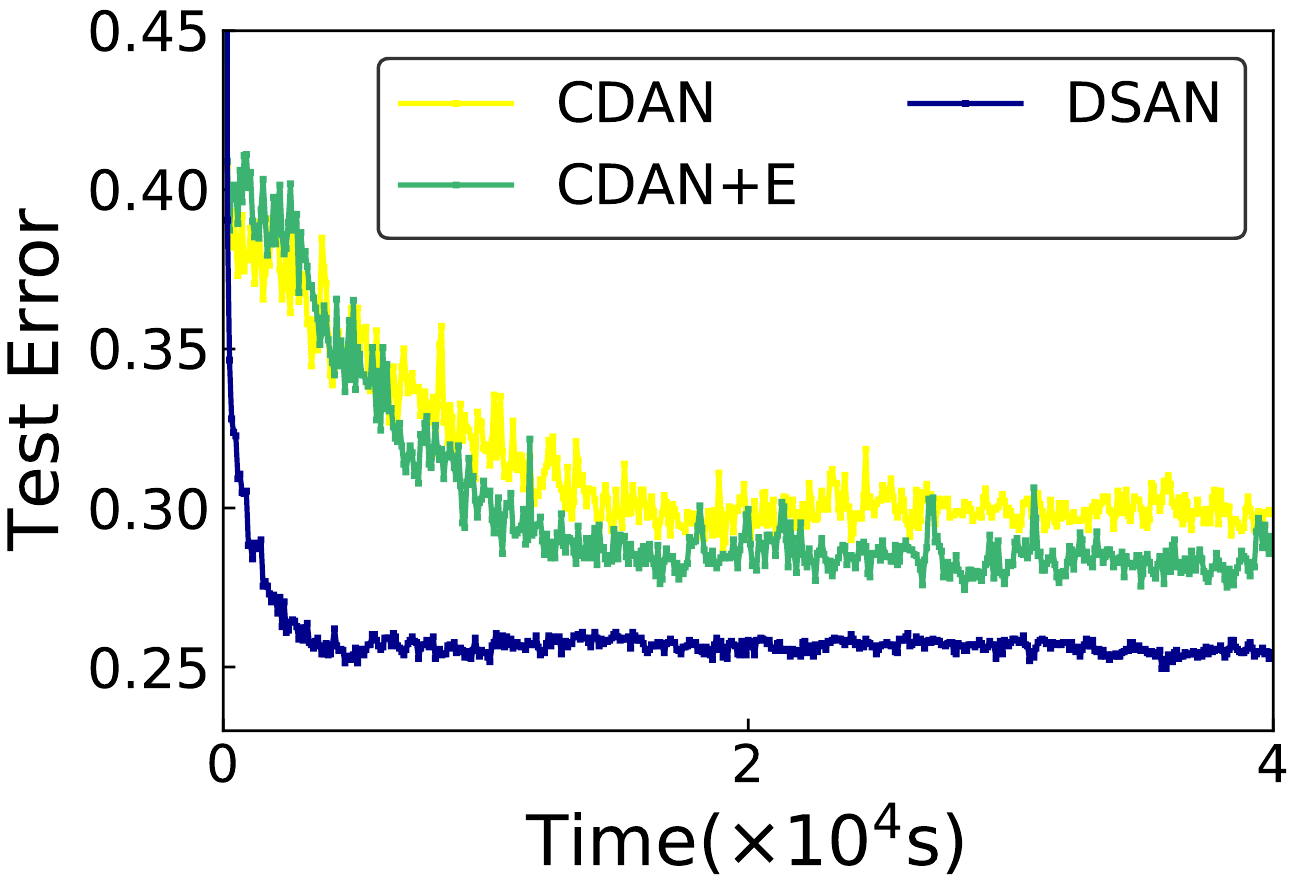}
\label{fig:convergenceb}
\end{minipage}
}
\caption{On task D $\rightarrow$ A (Office31), we further analyze the convergence. (a) and (b) are convergence about iterations and time.}
\label{fig:convergence}
\end{figure}

\textbf{Implementation Details} 
For object recognition tasks, we employed the ResNet~\cite{he2016deep}. Following CDAN~\cite{long2018conditional}, a bottleneck layer $fcb$ with 256 units is added after the last average pooling layer for safe transfer representation learning. We use the output of $fcb$ as inputs to the LMMD. Note that, it is easy to add LMMD in multiple layers and we only add LMMD to one layer. Also, image random flipping and cropping are adopted following JAN~\cite{long2016deep}. For a fair comparison, all baselines use the same architecture (For VisDA-2017, we use ResNet101~\cite{he2016deep}, while ResNet50 for others). We fine-tune all convolutional and pooling layers from ImageNet pre-trained models and train the classifier layer via back-propagation. Since the classifier is trained from scratch, we set its learning rate to be 10 times that of the other layers. For digit classification tasks, we follow the protocols in ADDA~\cite{tzeng2017adversarial}, and use the same architecture with ADDA. 

For all tasks, we use mini-batch stochastic gradient descent (SGD) with momentum of 0.9 and the learning rate annealing strategy in Revgrad~\cite{ganin2015unsupervised}: the learning rate is not selected by a grid search due to high computational cost, it is adjusted during SGD using the following formula: ${\eta}_\theta = \frac{\eta_0}{(1+\alpha \theta)^\beta}$, where $\theta$ is the training progress linearly changing from $0$ to $1$, $\eta_0 = 0.01$, $\alpha = 10$ and $\beta = 0.75$. To suppress noisy activations at the early stages of training, instead of fixing the adaptation factor $\lambda$, we gradually change it from $0$ to $1$ by a progressive schedule: $\lambda_\theta = \frac{2}{exp(-\gamma \theta)} - 1$, and $\gamma = 10$ is fixed throughout the experiments~\cite{ganin2015unsupervised}. 

We implement DSAN in PyTorch and report the average classification accuracy and standard error of three random trials. For all MMD-based methods~\cite{long2015learning,long2016deep,tzeng2014deep} including DSAN, we adopt Gaussian kernel with bandwidth set to median pairwise squared distances on the training data~\cite{gretton2012kernel}.

\subsection{Results}

\textbf{Object Recognition} The classification results of ImageCLEF-DA, Office-31, Office-Home, VisDA-2017, and Adaptiope are respectively shown in Table~\ref{tab:Image-CLEF},~\ref{tab:office31},~\ref{tab:officehome},~\ref{tab:visda}, and~\ref{tab:adaptiope}. DSAN outperforms all compared methods on most transfer tasks. In particular, DSAN substantially improves the average accuracy by large margins (more than 3\%) on Image-CLEF, Office-Home, VisDA-2017 and Adaptiope. The encouraging results indicate the importance of Subdomain Adaptation and show that DSAN is able to learn more transferable representations.

The experimental results further reveal several insightful observations. \textbf{(1)} In standard domain adaptation, Subdomain Adaptation methods (MADA~\cite{pei2018multi}, CDAN~\cite{long2018conditional}, our DSAN) outperform previous global domain adaptation methods. The improvement from previous global domain adaptation methods to Subdomain Adaptation methods is crucial for domain adaptation: previous methods align global distribution without considering the relationship between subdomains, while DSAN accurately aligns the relevant subdomain distributions which can capture more fine-grained information for each category. \textbf{(2)} In particular, comparing DSAN with the most recent Subdomain Adaptation methods~\cite{long2018conditional,pei2018multi}, DSAN achieves better performance. This verifies the effectiveness of our model. \textbf{(3)} Comparing DSAN with the non-adversarial methods~\cite{long2015learning,long2016deep,sun2016deep,tzeng2014deep}, DSAN also largely improves the average performance on 24 object recognition tasks (6.65\% higher than JAN~\cite{long2016deep}). \textbf{(4)} Comparing DSAN with LMMD, DAN with MMD and JAN with JMMD, DSAN achieves the best performance, which implies that LMMD is more suitable for aligning distributions than MMD and JMMD.

\textbf{Digit Classification} The classification results of three tasks of MNIST-USPS-SVHN are shown in Table~\ref{tab:digit}. Except for DRCN~\cite{ghifary2016deep}, all other baselines are adversarial ones. DSAN largely outperforms all baselines except SVHN $\rightarrow$ MNIST task. Comparing DSAN with MSTN~\cite{xie2018learning} which is also a Subdomain Adaptation method, DSAN achieves better average accuracy and more stable results with lower standard error.

Overall, all the above results demonstrate the effectiveness of the proposed model.

\subsection{Analysis}

\textbf{Feature Visualization:} We visualize in Figure~\ref{fig:3a}-~\ref{fig:3b} the network activations of task \textbf{A} $\rightarrow$ \textbf{W} learned by JAN and DSAN (both use Gaussian kernel) using t-SNE embeddings~\cite{donahue2014decaf}. Red points are source samples and blue are target samples. Figure~\ref{fig:3a} shows the result for JAN~\cite{long2016deep}, which is a typical statistic moment matching based approach using JMMD. We can find that the source and target domains are not aligned very well and some points are hard to classify. In contrast, Figure~\ref{fig:3b} shows the representations learned by our DSAN using LMMD. It is observed that source and target domains are aligned very well. We not only can see that the subdomains in different domains with the same class are very close, but also the subdomains with different classes are dispersed. This result suggests that our model DSAN is able to capture more fine-grained information for each category than JAN, and LMMD is more effective than JMMD to align the distributions.

\textbf{Distribution Discrepancy:} We use $\mathcal{A}$-distance and $\mathcal{A}_L$-distance mentioned at Sec.~\ref{A-distance} to measure global distribution discrepancy and the subdomain distribution discrepancy. Figure~\ref{fig:3c} shows $d_\mathcal{A}$ and $d_{\mathcal{A}_L}$ on task \textbf{A} $\rightarrow$ \textbf{W} with representations of CNN, JAN and DSAN. We observe that $d_\mathcal{A}$ and $d_{\mathcal{A}_L}$ using DSAN are much smaller than the ones using CNN and JAN, which shows that DSAN can not only close the cross-domain gap but also one of relevant subdomains more effectively.

MMD is a method to measure the discrepancy of global distributions, while LMMD is a method to measure the discrepancy of local subdomain distributions. We compute MMD and LMMD across domains on task \textbf{A} $\rightarrow$ \textbf{W} using CNN, JAN and DSAN, based on the features in $pool$ layer and ground-truth labels. Figure~\ref{fig:3d} shows that both MMD and LMMD using DSAN activations are much smaller than using CNN and JAN activations, which again validates that DSAN successfully reduces the discrepancy of global and local distributions. In addition, LMMD is smaller than MMD for the reason that LMMD can estimate the distribution discrepancy by eliminating the irrelevant data.

\textbf{Convergence}: We testify the convergence of CDAN, CDAN+E, and DSAN, with the test errors on task D $\rightarrow$ A (Office31) shown in Figure~\ref{fig:convergence}. From Figure~\ref{fig:convergencea}, with the same number of iterations, DSAN achieves faster convergence than CDAN and CDAN+E. From Figure~\ref{fig:convergenceb}, with the same period of time, DSAN also converges faster. Besides, the results further reveal that for each iteration DSAN runs faster than CDAN and CDAN+E.

\begin{table}[!th]
	\centering
	\caption{The comparison of the Subdomain Adaptation methods. $K$ in MADA means the number of classes. Parameter means the number of hyperparameters in the methods. Time means the average convergence time on the ImageCLEF-DA dataset (seconds). Time is measured on a GeForce GTX 1080 Ti GPU by ourselves. Accuracy means the average accuracy on the ImageCLEF-DA dataset. '-' means we dose not find the results from the original paper.} \label{tab:compare}
	\begin{tabular}{@{}cccccc@{}}
		\toprule
		Method & MADA & MSTN & CDAN & Co-DA & \textbf{DSAN} \\
		\midrule
		Adversarial & Yes & Yes & Yes & Yes & \textbf{No} \\
		Loss terms & $1+K$ & 3 & 3 & 12 & \textbf{2} \\
		Parameter & 1 & 3 & 1 & 6 & \textbf{1} \\
		Time(s) & 4318 & - & 1944 & - & \textbf{702} \\
		Accuracy & 85.8 & - & 87.1 & - & \textbf{90.2} \\
		\bottomrule
	\end{tabular}
\end{table}

\textbf{Discussion on the Advantage of DSAN:}
To give an overview of the results, we further compare our DSAN with several adversarial subdomain adaptation methods~\cite{kumar2018co,pei2018multi,xie2018learning,long2018conditional} in Table~\ref{tab:compare}, and find some insightful observations. First, the adversarial Subdomain Adaptation methods usually have several loss functions, while DSAN only needs one classification loss and one LMMD loss. In addition, DSAN only has one hyperparameter, while MSTN~\cite{xie2018learning} and Co-DA~\cite{kumar2018co} have several hyperparameters. DSAN has fewer loss terms and hyperparameter, which also indicates the easy implementation. Second, Comparing DSAN with MADA~\cite{pei2018multi} and CDAN~\cite{long2018conditional}, DSAN also takes less time to converge. Third, DSAN achieves the best performance. Especially, DSAN achieves \textbf{3\%} accuracy higher than CDAN~\cite{long2018conditional} which is one of the most recent Subdomain Adaptation methods. Overall, all the results again validate the advantage of our model DSAN.

\section{Conclusion}
\label{sec:conclusion}
Unlike the previous methods that align the global source and target distributions, \textit{Subdomain Adaptation} can accurately align the distributions of relevant subdomains within the same category of the source and target domains. However, most recent subdomain adaptation methods are adversarial approaches which contain several loss functions and converge slowly. Based on this, we proposed a new method Deep Subdomain Adaptation Network (DSAN), which is a non-adversarial method and very simple and easy to implement. Furthermore, to measure the discrepancy between relevant subdomains within the same category of different domains, we proposed a new local distribution discrepancy measure LMMD. Extensive experiments conducted on both object recognition and digit classification tasks demonstrate the effectiveness of the proposed model.
\section{Acknowledgments}
\label{sec:ack}
The research work supported by the National Key Research and Development Program of China under Grant No. 2018YFB1004300, the National Natural Science Foundation of China under Grant No. U1836206, U1811461, 61773361, the Project of Youth Innovation Promotion Association CAS under Grant No. 2017146.
\ifCLASSOPTIONcaptionsoff
  \newpage
\fi



%
\bibliographystyle{IEEEtran}
\bibliography{bare_jrnl.bbl}

%



\begin{IEEEbiography}[{\includegraphics[width=1in,height=1.25in,clip,keepaspectratio]{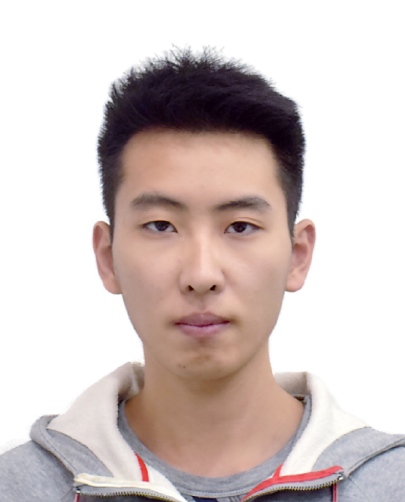}}]{Yongchun Zhu}
is currently pursuing his M.S. degree in the Institute of Computing Technology, Chinese Academy of Sciences, Beijing, China. He has published some papers in journals and conference proceedings including Neural Networks, AAAI, WWW, and PAKDD. He received his B.S. degree from Beijing Normal University, China in 2018. His main research interests include transfer learning, meta learning and recommendation system.
\end{IEEEbiography}

\begin{IEEEbiography}[{\includegraphics[width=1in,height=1.25in,clip,keepaspectratio]{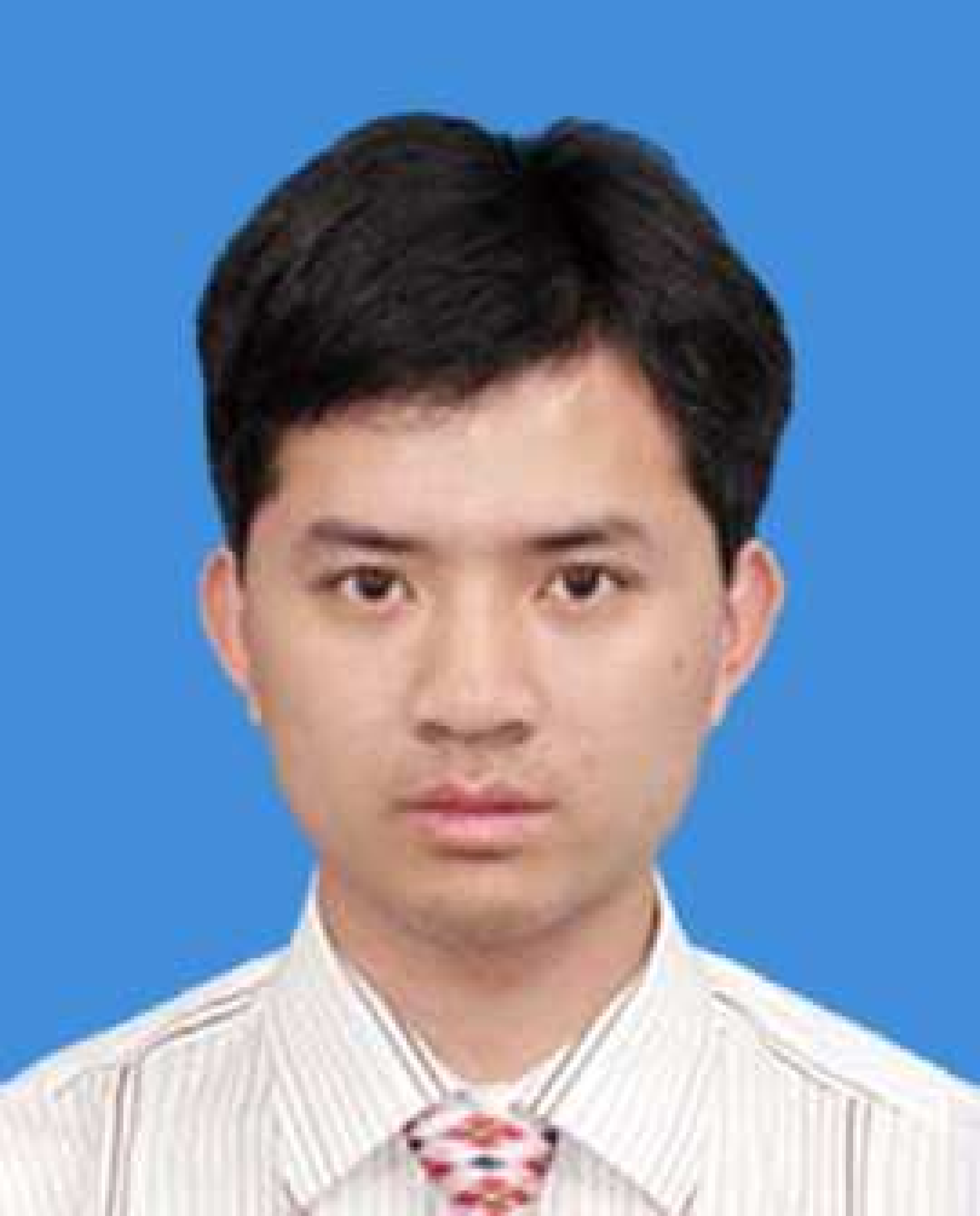}}]{Fuzhen Zhuang}
is an associate professor in the Institute of Computing Technology, Chinese Academy of Sciences. His research interests include transfer learning, machine learning, data mining, multi-task learning and recommendation systems. He has published more than 100 papers in the prestigious refereed journals and conference proceedings, such as IEEE TKDE, IEEE Transactions on Cybernetics, IEEE TNNLS, ACM TIST, SIGKDD, IJCAI, AAAI, WWW, and ICDE. 
\end{IEEEbiography}

\begin{IEEEbiography}[{\includegraphics[width=1in,height=1.25in,clip,keepaspectratio]{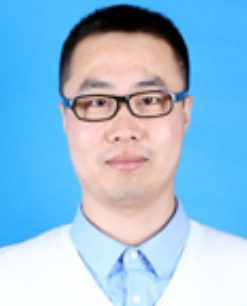}}]{Jindong Wang}
is currently a researcher as Microsoft Research, Beijing, China. He received his Ph.D degree from Institute of Computing Technology, Chinese Academy of Sciences, Beijing, China. His research interest mainly includes transfer learning, machine learning, data mining, and artificial intelligence. He serves as the reviewers of many journals and conference, including TPAMI, CSUR, ACM CHI, and Neurocomputing. He is also a member of IEEE, ACM, AAAI, and CCF.
\end{IEEEbiography}
\vspace{-0.5cm}

\begin{IEEEbiography}[{\includegraphics[width=1in,height=1.25in,clip,keepaspectratio]{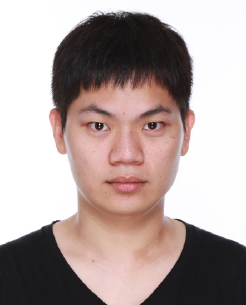}}]{Guolin Ke}
is currently a Senior Researcher in Machine Learning Group, Microsoft Research Asia. His research interests mainly lie in machine learning algorithms. He created one of the most popular decision tree learning tool LightGBM.
\end{IEEEbiography}
\vspace{-0.5cm}

\begin{IEEEbiography}[{\includegraphics[width=1in,height=1.25in,clip,keepaspectratio]{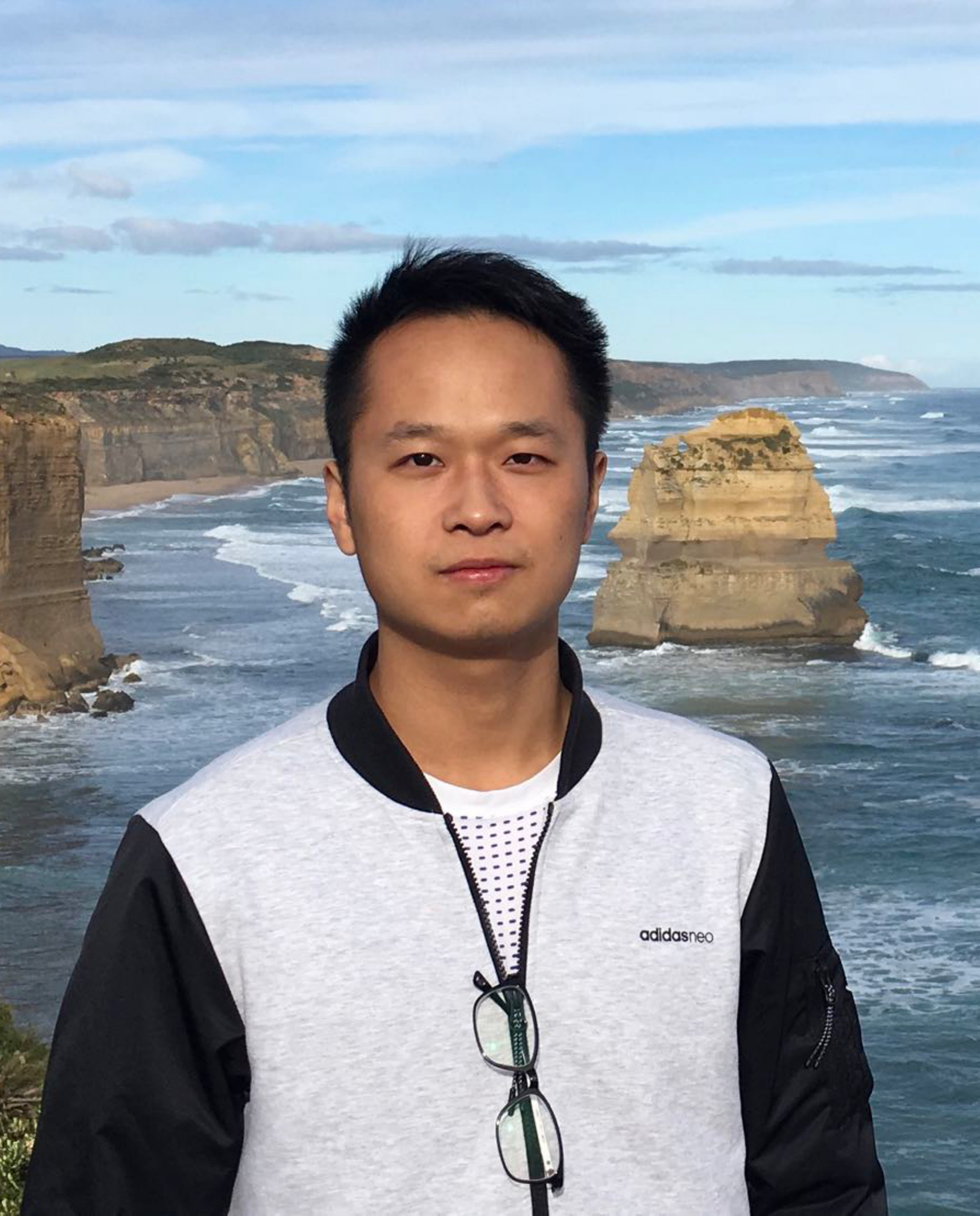}}]{Jingwu Chen}
received the master’s degree in Institute of Computing Technology, Chinese Academy of Sciences, in 2019. He is currently a  member of  Recommendation Team in ByteDance. His research interests include machine learning and its applications, such as recommendation system and computational advertising.
\end{IEEEbiography}
\vspace{-0.5cm}

\begin{IEEEbiography}[{\includegraphics[width=1in,height=1.25in,clip,keepaspectratio]{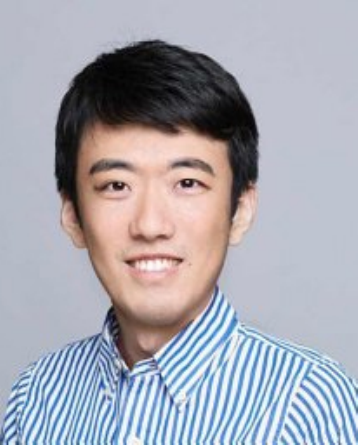}}]{Jiang Bian}
is a researcher and engineer with rich experience in information retrieval, data mining and machine learning. Now, he is a Principal Researcher and Research Manager at Microsoft Research with research interests in AI for finance, AI for logistics, deep learning, multi-agent reinforcement learning, computational advertising, and a variety of machine learning applications. Prior to that, he was a Senior Scientist in Yidian Inc., a startup company in China, where he has been working on recommendation and search problems.
\end{IEEEbiography}
\vspace{-0.5cm}

\begin{IEEEbiography}[{\includegraphics[width=1in,height=1.25in,clip,keepaspectratio]{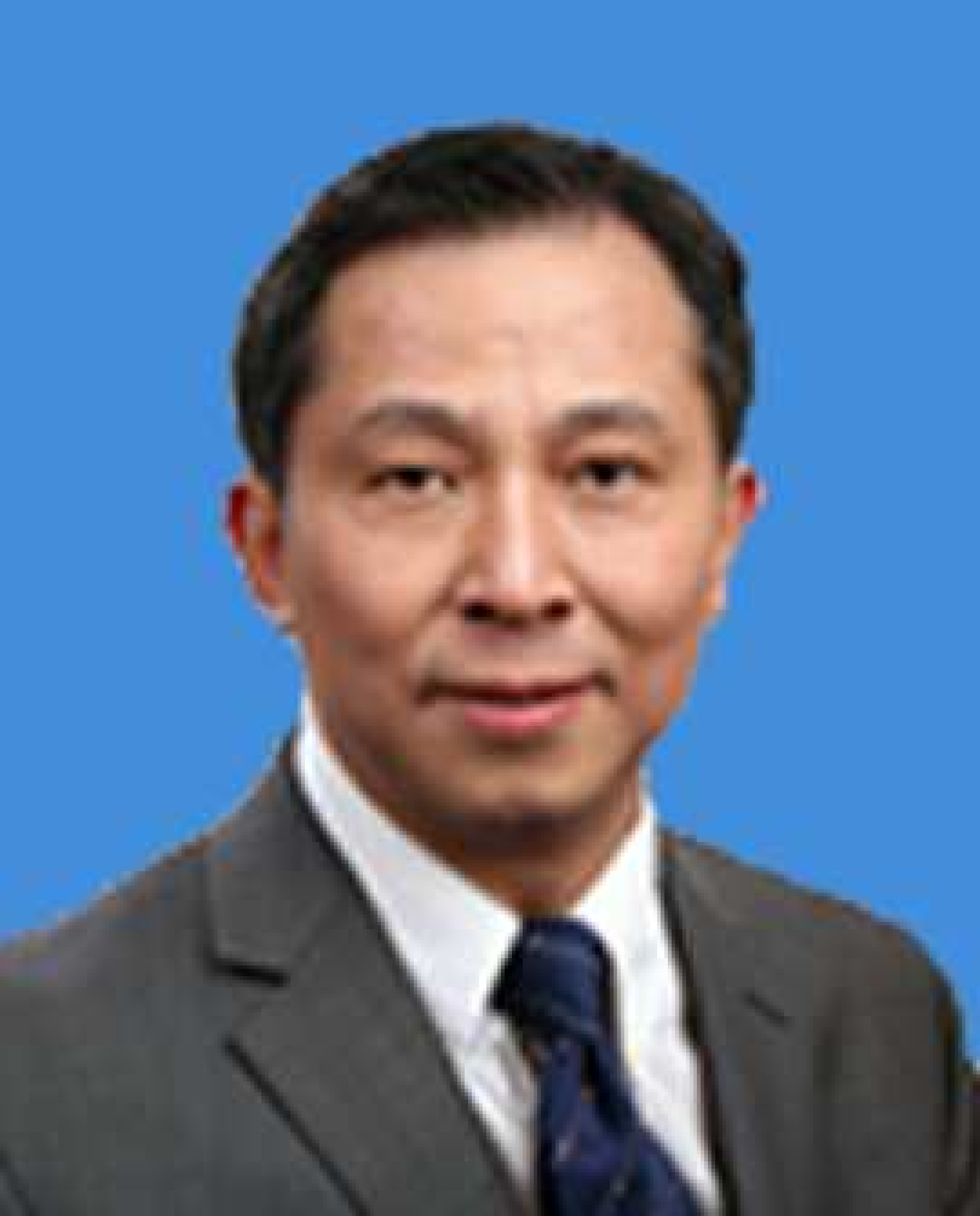}}]{Hui Xiong}
is currently a Full Professor at Rut- gers University, where he received the 2018 Ram Charan Management Practice Award as the Grand Prix winner from the Harvard Business Review, RBS Deans Research Professorship (2016), the Rutgers University Board of Trustees Research Fellowship for Scholarly Excellence (2009), the IEEE ICDM Best Research Paper Award (2011), and the IEEE ICDM Outstanding Service Award (2017). He received the Ph.D. degree in Computer Science from the University of Minnesota - Twin Cities, USA, in 2005. He is a co-Editor-in-Chief of Encyclopedia of GIS, an Associate Editor of IEEE TBD, ACM TKDD, and ACM TMIS. He has served regularly on the organization committees of numerous conferences, such as a Program Co-Chair for ACM KDD 2018 (research track), ACM KDD 2012 (industry track), IEEE ICDM 2013, and a General Co-Chair for IEEE ICDM 2015. He is an IEEE Fellow and an ACM Distinguished Scientist.
\end{IEEEbiography}
\vspace{-0.5cm}

\begin{IEEEbiography}[{\includegraphics[width=1in,height=1.25in,clip,keepaspectratio]{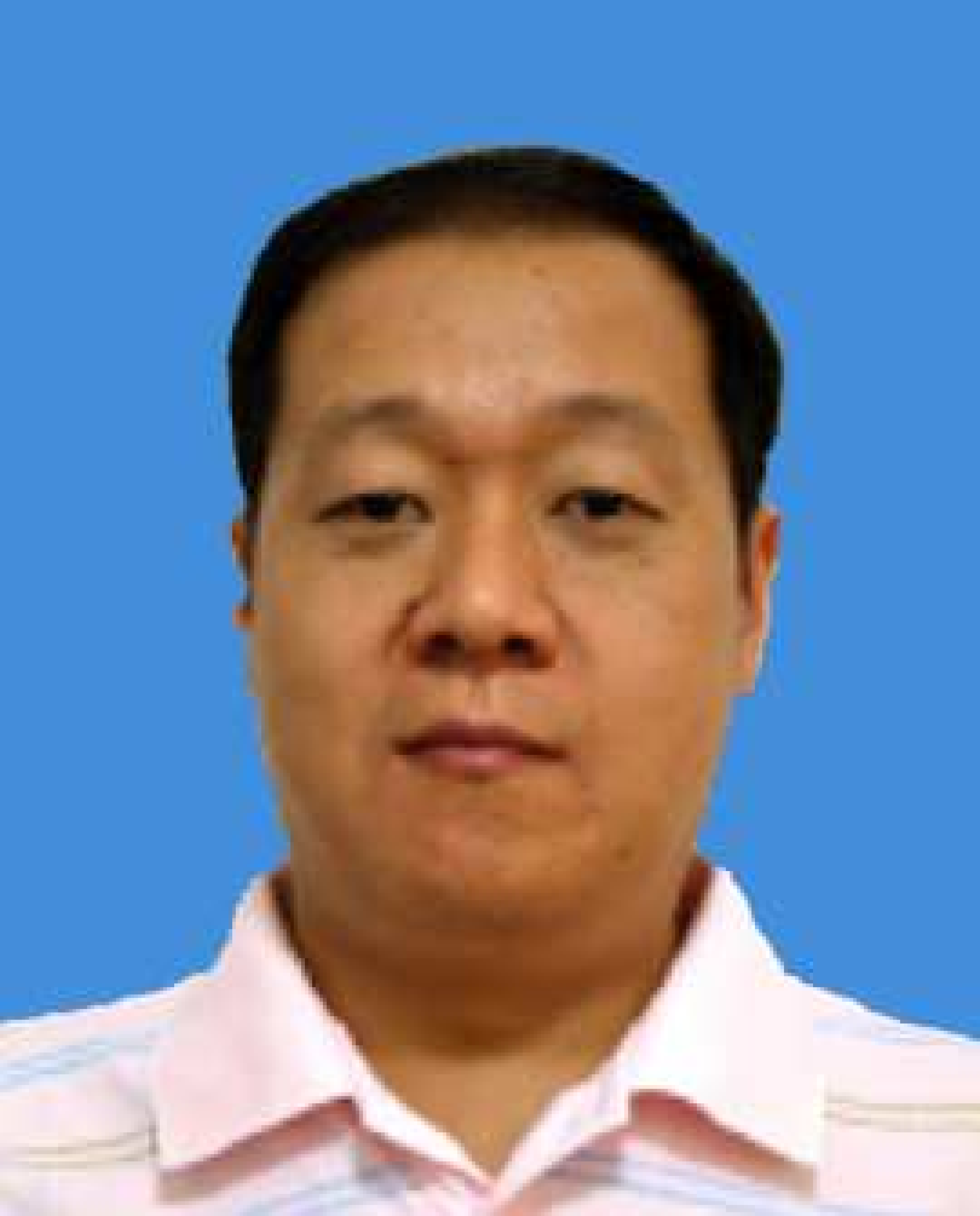}}]{Qing He}
is a Professor in the Institute of Computing Technology, Chinese Academy of Science (CAS), and he is a Professor at the Graduate University of Chinese (GUCAS). He received the B.S degree from Hebei Normal University, Shijiazhang, P. R. C., in 1985, and the M.S. degree from Zhengzhou University, Zhengzhou, P. R. C., in 1987, both in mathematics. He received the Ph.D. degree in 2000 from Beijing Normal University in fuzzy mathematics and artificial intelligence, Beijing, P. R. C. Since 1987 to 1997, he has been with Hebei University of Science and Technology. He is currently a doctoral tutor at the Institute of Computing and Technology, CAS. His interests include data mining, machine learning, classification, fuzzy clustering. 
\end{IEEEbiography}






\end{document}